\documentclass[runningheads]{llncs}

\usepackage{eccv}

\usepackage{eccvabbrv}

\usepackage{graphicx}
\usepackage{booktabs}

\usepackage[accsupp]{axessibility}  

\usepackage{bm}
\usepackage{xcolor}
\usepackage{multirow}
\usepackage{makecell}
\usepackage{textcomp}
\usepackage{graphicx}
\usepackage{amssymb} 
\usepackage{amsmath}
\usepackage{comment}
\usepackage{wrapfig}
\usepackage{newfloat}
\usepackage{caption}
\usepackage{tablefootnote}
\usepackage{booktabs}
\usepackage{inconsolata}
\usepackage{enumitem}
\usepackage{algpseudocode}
\usepackage{algorithm}
\usepackage{colortbl}
\usepackage{graphicx}
\usepackage{booktabs}

\usepackage{hyperref}

\usepackage{orcidlink}

\begin{document}

\title{BI-MDRG: Bridging Image History in Multimodal Dialogue Response Generation} 

\titlerunning{Briding Image History in Multimodal Dialogue Response Generation}

\author{Hee Suk Yoon\inst{1}\thanks{Equal Contribution}\orcidlink{0000-0003-2115-8459} \and
Eunseop Yoon\inst{1}$^\star$\orcidlink{0000-0002-5580-5354} \and
Joshua Tian Jin Tee\inst{1}$^\star$\orcidlink{0009-0001-5119-2802} \and
Kang Zhang\inst{1}\orcidlink{0000-0003-2761-9383} \and
Yu-Jung Heo\inst{2}\orcidlink{0000-0002-5725-9545} \and
Du-Seong Chang\inst{2}\orcidlink{0000-0001-6692-962X} \and
Chang D. Yoo\inst{1}\thanks{Corresponding Author}\orcidlink{0000-0002-0756-7179}
}

\authorrunning{HS.~Yoon et al.}

\institute{Korea Advanced Institute of Science and Technology (KAIST) \and
KT Corporation \\
\email{\{hskyoon,esyoon97,joshuateetj,zhangkang,cd\_yoo\}@kaist.ac.kr} \\
\email{\{yj.heo,dschang\}@kt.com}
}

\maketitle

\begin{abstract}
\textit{Multimodal Dialogue Response Generation (MDRG)} is a recently proposed task where the model needs to generate responses in texts, images, or a blend of both based on the dialogue context. Due to the lack of a large-scale dataset specifically for this task and the benefits of leveraging powerful pre-trained models, previous work relies on the text modality as an intermediary step for both the image input and output of the model rather than adopting an end-to-end approach.
However, this approach can overlook crucial information about the image, hindering 1) image-grounded text response and 2) consistency of objects in the image response. In this paper, we propose \textbf{BI-MDRG} that bridges the response generation path such that the image history information is utilized for enhanced relevance of text responses to the image content and the consistency of objects in sequential image responses. Through extensive experiments on the multimodal dialogue benchmark dataset, we show that BI-MDRG can effectively increase the quality of multimodal dialogue. Additionally, recognizing the gap in benchmark datasets for evaluating the image consistency in multimodal dialogue, we have created a curated set of 300 dialogues annotated to track object consistency across conversations. The code and the dataset is publicly available at \href{https://github.com/hee-suk-yoon/BI-MDRG}{https://github.com/hee-suk-yoon/BI-MDRG}.  
\end{abstract}

\section{Introduction}
\label{sec:intro}
With the development of instant messaging technology, visual modalities are increasingly used alongside text in online communication. To enhance user interaction with intelligent agents, a new task, Multimodal Dialogue Response Generation (MDRG) \cite{divter}, has been proposed. This task requires models to generate both text and image responses based on dialogue history containing texts and images.
\begin{figure}[t!]
	\centering
	\includegraphics[width=1.0\linewidth]{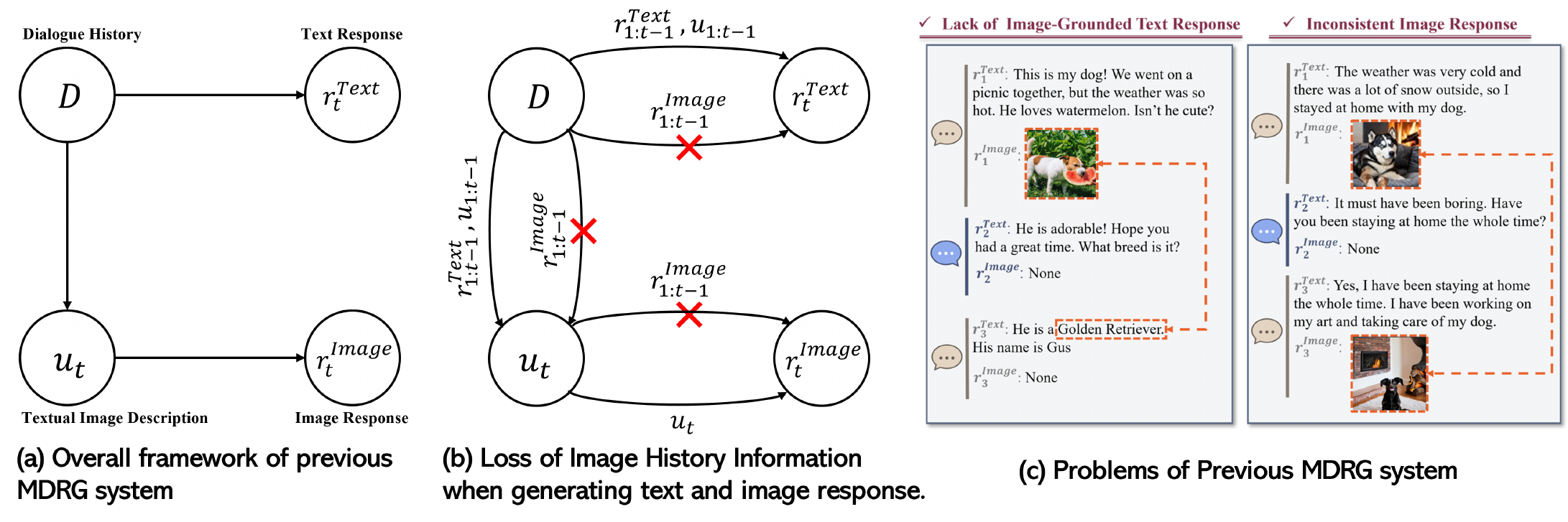}
	\caption{(a) Outlines the framework of previous Multimodal Dialogue Response Generation (MDRG) systems, which uses the textual descriptions of images ($u_t$) as an intermediary step toward generating image responses (\(r_{t}^{\text{Image}}\)). (b) Highlights the limitations of these systems, particularly their failure to fully leverage image history (\(r_{1:t-1}^{\text{Image}}\)) in crafting both the textual response ($r_{t}^{\text{Text}}$) and the image response ($r_{t}^{\text{Image}}$). (c) Illustrates the consequences of this oversight, including responses that lack grounding in image context and consistency in image-based replies.}
	\label{fig:1}
\end{figure}
Since learning an effective multimodal generation model with a single sequence-to-sequence model requires a large number of training instances, the lack of a large-scale multimodal dialogue dataset poses a significant challenge for the task. 

To solve this issue, the previous MDRG model~\cite{divter} disentangles the textual response and image response generation, as illustrated in Figure \ref{fig:1}-(a). Specifically, given the dialogue history $D$ at $t$-th conversation turn, the MDRG model first generates the textual response $r^\text{Text}_t$ and the intermediary textual image description $u_t$. This image description $u_t$ is subsequently fed into a text-to-image model to generate the corresponding image $r^\text{Image}_t$. Such an approach allows the utilization of powerful models pre-trained on vast amounts of available data for text-to-text and text-to-image pairs, bypassing the need for a large-scale multimodal dialogue dataset for direct end-to-end training.

Nonetheless, due to its reliance on text as an intermediary representation of images, the previous MDRG model overlooks crucial information about the image, as shown in Figure \ref{fig:1}-(b). By converting images from the dialogue history into textual descriptions (\(u_{1:t-1}\)), these models fail to fully utilize the rich visual content in the actual images (\(r_{1:t-1}^{\text{Image}}\)). This leads to two major issues: a lack of image-grounded context in textual responses (Figure \ref{fig:1}-(c) \textit{(left)}) and inconsistencies in image generation across dialogues (Figure \ref{fig:1}-(c) \textit{(right)}).

For instance, Figure \ref{fig:1}-(c) \textit{(left)} shows the existing MDRG model failing to provide an image-grounded response to `What breed is your dog?' because it only perceives the dog through the text `a dog eating a watermelon'. Figure \ref{fig:1}-(c) \textit{(right)} illustrates the model's inconsistency, where the `dog' in the image history is not maintained in the generated response. These issues highlight the need for improved MDRG models that effectively utilize image history.

\paragraph{\textbf{Contribution}} This paper introduces BI-MDRG to enhance text and image responses by bridging the image history.
\begin{itemize}
\setlength\itemsep{0em}
\item[$\bullet$] \textit{Bridging Image History in Text Responses:} In Section \ref{3.1}, we propose an architectural modification in which the visual features, extracted using a visual encoder, are integrated into the cross-attention layers of the core language model. This is complemented by a novel multimodal causal attention mask modulation tailored for the MDRG task to improve image-grounded textual responses.

\item[$\bullet$] \textit{Bridging Image History in Image Responses:} 
In Section \ref{3.2}, we propose a citation framework where we use the Citation Module designed to augment textual image descriptions by tagging objects with citation tags that identify recurring objects throughout a dialogue. In Section \ref{3.3}, we show that training with these augmented data enables the model to recognize and maintain the consistency of objects in subsequent image responses during inference using Customized Text-to-Image Models. Due to the absence of benchmark datasets for evaluating image consistency in multimodal dialogues, we have created the Multimodal Dialogue Image Consistency (MDIC) dataset containing dialogues annotated to track object consistency across conversations.
\end{itemize}

\section{Related Work}

\subsection{Multimodal Dialogue Datasets}
\label{related_works}
Multimodal dialogue datasets generally fall into three categories: question and answering \cite{related_visdialog, related_avsd} (where the task involves asking and answering questions about a specific image), in-scene \cite{related_openvidial, related_openvidial2, related_ytd, related_meld, related_m3ed} (where each dialogue turn corresponds to a scene from a movie or video), and conversation-based \cite{imagechat, related_photochat, related_mmdd, related_dialogcc, mmdialog,related_tiktalk} (which engage in natural dialogue about a given image or involves image sharing within natural conversations). Further details are in Appendix C.1.

This paper primarily explores the model's capability for natural dialogue within the conversation-based category. Notable datasets in this segment include ImageChat \cite{imagechat}, PhotoChat \cite{related_photochat}, MMDD \cite{related_mmdd}, DialogCC \cite{related_dialogcc}, MMChat \cite{related_mmchat}, TikTalk \cite{related_tiktalk}, and MMDialog \cite{mmdialog}. Given our focus on English-language scenarios, we exclude MMChat and Tiktalk from our evaluation, as it is a dataset primarily in Chinese. Additionally, DialogCC (not publicly available) and MMDD, which are synthesized by algorithmically pairing images with text-only dialogues for random turns, are also excluded from our analysis. Therefore, our evaluation is centered on ImageChat, PhotoChat, and MMDialog.

ImageChat \cite{imagechat} consists of image-centered dialogues, where each dialogue is centered around a single given image. PhotoChat \cite{related_photochat} features dialogues collected from social media, where a single image is shared in one of the conversation turns, which better mirrors everyday human interaction. Still, their limited scale and domain diversity restrict their applicability. Overcoming these limitations, MMDialog \cite{mmdialog} features over a million diverse dialogues from social media, where multiple images are shared across numerous conversation turns, providing a more realistic representation of open-domain multimodal conversations.

\subsection{Multimodal Dialogue Modeling}
Pioneering studies \cite{related_modeling1,related_modeling2,related_modeling3} have delved into improving the performance of image-grounded conversational agents. While \cite{related_mmchat, yoon-etal-2022-information, yoon2023hear, kim2021structuredcoreferencegraphattention} introduced a Seq2Seq based model focusing on multimodal dialogues, it primarily generated textual responses, not fully embracing the multimodal response scenario. In a notable advancement, \cite{divter} presented Divter, which not only produces informative text but also generates high-resolution images, marking a significant leap forward in multimodal dialogue response generation (MDRG). \textit{It is important to note that our focus in this paper is generation-based models, contrasting with retrieval-based \cite{retrieval_1, retrieval_2, retrieval_3}, which output image responses by retrieving existing images from a corpus instead of generating new ones.}
\subsection{Customized Text-to-Image}
Recent studies on text-to-image diffusion models \cite{stablediffusion,diffusion1,diffusion2, koo2024wavelet} focus on customization \cite{customize1,customize2,customize3,customize4}, learning specific concepts from a few images. Following this, users can flexibly generate the learned concepts into new scenes. Text inversion \cite{customize2} generates varied contexts for a concept by updating the text embedding without altering the model. Dreambooth \cite{customize4} and Custom Diffusion \cite{customize3} fine-tune the U-Net architecture using an identifier, class label, and images. In a notable enhancement, BLIP-Diffusion \cite{blipdiffusion} enables zero-shot subject-driven generation, allowing fast customized text-to-image generation.

\section{BI-MDRG: Bridging the Image History in Multimodal Dialogue Response Generation}
We introduce BI-MDRG, a conversational agent designed to produce both textual and visual responses with enhanced awareness of image history. Sections \ref{3.1} and \ref{3.2} detail the training procedure, effectively integrating image history information into text responses and textual image descriptions. Section \ref{3.3} outlines the inference process, wherein the image history informs the image responses by leveraging the captured details from enhanced textual image descriptions. 

\subsection{Bridging the Image History for Image-Grounded Text Response}
\label{3.1}
\begin{wrapfigure}{r}{0.35\textwidth}
\vspace{-25pt}
	\centering
	\includegraphics[width=0.75\linewidth]{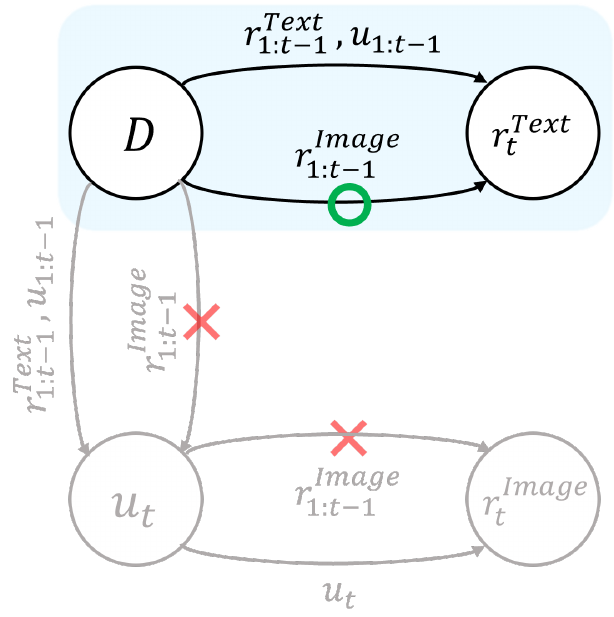}
	\caption{Bridging Image History to the Text Response.}
	\label{fig:method1}
\vspace{-30pt}
\end{wrapfigure}
The example shown in Figure \ref{fig:1}-(c) \textit{(left)} highlighted one of the crucial limitations in the previous MDRG system: their reliance on textual descriptions for understanding image history, which hinders the image-grounded textual responses. To overcome this, we have adopted an architectural change along with a multimodal causal attention mask modulation to effectively bridge the image history information $r_{1:t-1}^{\text{Image}}$ to the text response $r_{t}^{\text{Text}}$ (as depicted in Figure \ref{fig:method1}).

\begin{figure}[t!]
	\centering
	\includegraphics[width=0.90\linewidth]{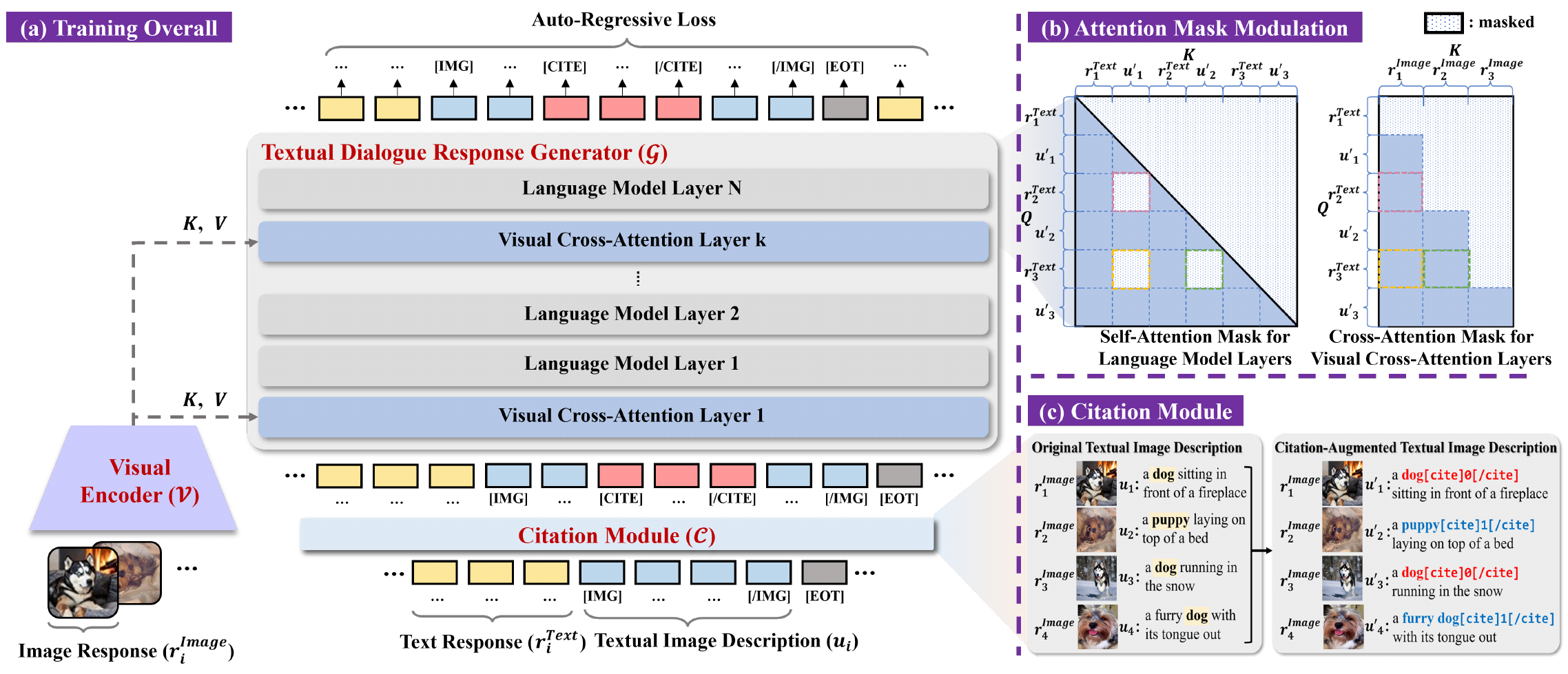}
	\caption{\textbf{Training of BI-MDRG.} (a) Textual Dialogue Response Generator $\mathcal{G}$ cross-attends to the image features from the Visual Encoder $\mathcal{V}.$ (b) Attention Mask Modulation alters the causal attention to prioritize image features over textual image descriptions. (c) Citation Module $\mathcal{C}$ generates citation-augmented textual image descriptions, enabling the tracking of objects within image history for consistency maintenance.} 
	\label{fig:model}
\end{figure}

\subsubsection{Architecture}
Our Textual Dialogue Response Generator $\mathcal{G}$ (Figure \ref{fig:model}-(a)), consists of a decoder-only language model with added visual cross-attention layers. 
These layers directly engage with image features provided by the Visual Encoder $\mathcal{V}$, drawing inspiration from Flamingo \cite{flamingo}, to reduce dependence on textual image descriptions for perceiving images. 
In our framework, a dialogue context $D=\{(r_i^{\text{Text}}, r_i^{\text{Image}})\}_{i=1}^n$ comprises multiple turns, each with an associated text response $r_i^{\text{Text}}$ and an image response $r_i^{\text{Image}}$. An image captioning model produces textual descriptions $u_i$ for each image $r_i^{\text{Image}}$, which are then transformed into citation-augmented descriptions $u'_i$ as shown in Figure \ref{fig:model}-(c) and further detailed in Section \ref{3.2} (\textit{if there is no image response for a turn, $r_i^{\text{Image}}=\emptyset$ and $u'_i=\emptyset$}). These descriptions and the text responses $\{r_1^{\text{Text}},u'_1,\ldots,r_n^{\text{Text}},u'_n\}$ are fed into $\mathcal{G}$, while the images $\{r_1^{\text{Image}}, \ldots, r_n^{\text{Image}}\}$ are processed by $\mathcal{V}$ to extract image features which are fed to the cross-attention layers in $\mathcal{G}$. Although our model directly cross-attends to the inputted image features, we retain the textual descriptions $u'_{1:n}$ as essential inputs since we require the generation of textual description by our $\mathcal{G}$ model, which subsequently gets used by a text-to-image model for constructing the image response. Keeping the textual description inputs allows efficient teacher-forced next token prediction training. 

\subsubsection{Multimodal Causal Attention Mask Modulation} 
For the input sequence $\{r_1^{\text{Text}},u'_1,\ldots,r_n^{\text{Text}},u'_n\}$, we use a specialized mask along with the standard causal mask (Figure \ref{fig:model}-(b)). The traditional causal mask allows each text response $r_i^{\text{Text}}$ to access previous textual image descriptions $u'_{1:i-1}$, leading to reliance on textual information over visual context. Our masking strategy prevents $r_i^{\text{Text}}$ from accessing $u'_{1:i-1}$, redirecting focus to the actual image features of $r_{1:i-1}^{\text{Image}}$, ensuring text responses are grounded on raw image features.

\begin{figure*}[t!]
	\centering
	\includegraphics[width=1.0\linewidth]{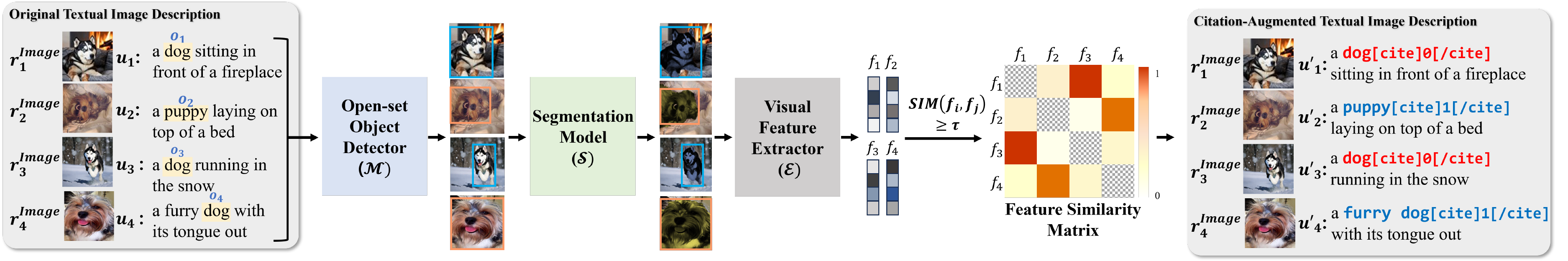}
	\caption{\textbf{Illustration of the Citation Module.} Citation Module recognizes identical objects within image history and injects this information into the textual image description with citation tags (e.g., [cite]0[/cite]).} 
	\label{fig:citation_example}
\end{figure*} 
\subsection{Citation Module: Bridging the Image History to the Textual Image Description}
\label{3.2}
\begin{wrapfigure}{r}{0.39\textwidth}
\vspace{-40pt}
	\centering
	\includegraphics[width=0.65\linewidth]{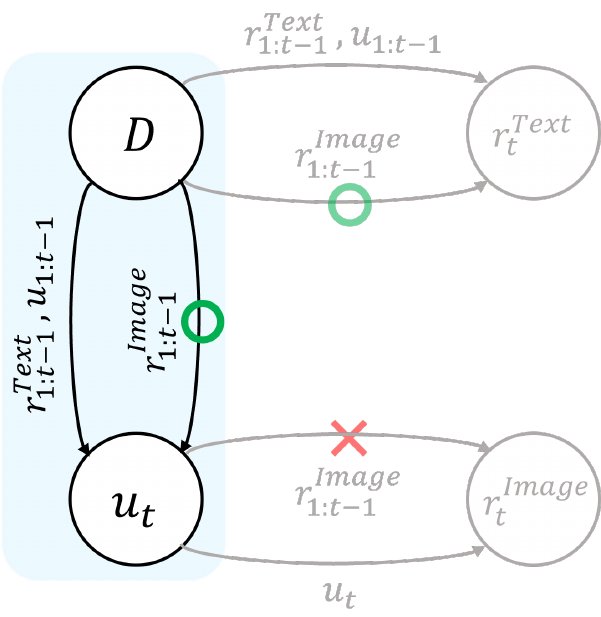}
	\caption{Bridging Image History to the Textual Image Description.}
	\label{fig:method2}
\vspace{-10pt}
\end{wrapfigure}
The example shown in Figure \ref{fig:1}-(c) \textit{(right)} highlighted another critical limitation of the previous MDRG system: its inability to ensure consistency in image responses. To address this, we propose the Citation Module (Figure \ref{fig:model}-(c)) to bridge the image history $r_{1:t-1}^{\text{Image}}$ and the textual image description $u_t$ (as depicted in Figure \ref{fig:method2}) by ensuring that the textual image description accurately relays which objects should persist in subsequent images. 

\subsubsection{Citation Module} 
Citation Module plays a pivotal role in tracking recurring objects in the dialogue using textual image descriptions. For instance, descriptions like "a dog is in front of a fireplace" and "a dog running in the snow" are augmented to "a dog\textbf{[cite]0[/cite]} is in front of a fireplace" and "a dog\textbf{[cite]0[/cite]} running in the snow," respectively if they reference the same dog.

Motivated by \cite{ma2023subject}, Figure \ref{fig:citation_example} details the citation process for textual image descriptions $\{u_1,...,u_n\}$. For each textual image description $u_i$, a Part of Speech (POS) tagging processor $\mathcal{P}$ is employed to tag words and pinpoint the key object word $o_i$ in the description. The word $o_i$, along with its corresponding image $r_i^{\text{Image}}$, is processed through an open-set object detector $\mathcal{M}$ to obtain the bounding box of the detected object, which is then input to a segmentation model $\mathcal{S}$ for generating object segmentation mask $s_i$. These masks are applied to isolate the objects from their backgrounds in $r_i^{\text{Image}}$ since background removal has been proven helpful for better extraction of object features \cite{chen2023anydoor}. These isolated objects are then analyzed by a visual feature extractor $\mathcal{E}$ to extract features $f_i$. The resulting feature set $\{f_1,...,f_n\}$ undergoes clustering based on cosine similarity to identify identical objects across images, as outlined in Algorithm \ref{alg:citation}; this involves assigning each element a cluster id $c_i$ based on the similarity of their features. For each $o_i$ in $u_i$, we augment the word so that it is followed by its corresponding cluster id (i.e., citation tag) $c_i$ of $f_i$, resulting in the citation augmented textual description $u'_i$. \textit{This Citation Module operates with off-the-shelf components for citation tags, requiring no training on the target dataset.}
\begin{algorithm}[t!]
\scriptsize
\caption{Citation Module}\label{alg:citation}
\begin{algorithmic}
\Require image response $(r_1^{\text{Image}}, ... r_n^{\text{Image}})$, textual image description $(u_1, ... u_n)$, similarity threshold $\tau$, POS tagging processor $\mathcal{P}$, open-set object detector $\mathcal{M}$, segmentation model $\mathcal{S}$, visual feature extractor $\mathcal{E}$, object cluster dictionary $\mathcal{K} = \{\}$
\For{$i \gets 1 \text{ to } n$}
    \State $o_i \gets$ $\mathcal{P}$($u_i$) \Comment{Identify principal object word in $u_i$.}
    \State $box_i \gets$ $\mathcal{M}$($o_i, r_i^{\text{Image}}$) \Comment{Obtain bounding box for $o_i$.}
    \State $s_i \gets$ $\mathcal{S}$($box_i$) \Comment{Obtain segmentation mask for $o_i$.}
    \State $f_i \gets$ $\mathcal{V}$($s_i \odot r_i^{\text{Image}}$) \Comment{Extract visual feature for $o_i$.}
\EndFor

\Statex

\State $cluster\_id \gets 0$
\For{$i \gets 1$ to $n$} \Comment{Perform clustering of $o_i$.}
    \If{not $\mathcal{K}$.has\_key($o_i$)} 
        \State $\mathcal{K}[o_i] \gets cluster\_id$ \Comment{Gets own cluster.}
        \State $cluster\_id \gets cluster\_id + 1$
    \EndIf
    \For{$j \gets i+1$ to $n$}
        \If{$\Call{Sim}{f_i, f_j} \geq \tau$ and not $\mathcal{K}$.has\_key($o_j$)}
            \State $\mathcal{K}[o_j] \gets \mathcal{K}[o_i]$ \Comment{Puts into existing cluster.}
        \EndIf
    \EndFor
\EndFor

\State \Return $\mathcal{K}$

\end{algorithmic}
\end{algorithm}
\subsubsection{Generative Training Objective} We use the next token prediction training via teacher forcing, which is used in standard auto-regressive language models \cite{gpt2}. Specifically, given the token sequence $w = \{w_j\}_{j=1}^{N}$ of the input sequence $\{r_1^{\text{Text}},u'_1,...,r_n^{\text{Text}},u'_n\}$ and the images $\{r_1^{\text{Image}},...,r_n^{\text{Image}}\}$, we minimize the negative log-likelihood: 
\begin{equation}
\mathcal{L}(w) = \sum_{j=1}^N \log P(w_j | w_{<j}, \{r_i^{\text{Image}}\}_{i=1}^n ; \mathcal{G},\mathcal{V}).
\label{eq:training}
\end{equation}
\textit{With such training, our model can generate textual image descriptions during inference with citation tags that reflect the objects needing consistency.}

\subsection{Inference Procedure: Bridging the Image History for Consistent Image Response}
\label{3.3}
\begin{wrapfigure}{r}{0.39\textwidth}
\vspace{-35pt}
	\centering
	\includegraphics[width=0.80\linewidth]{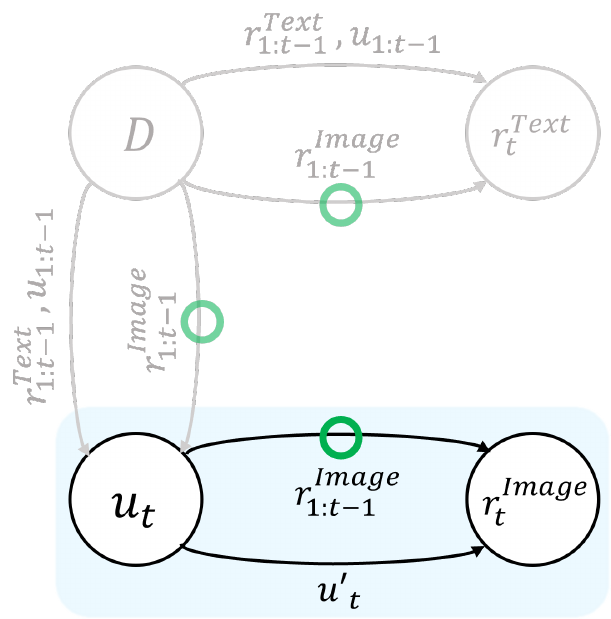}
	\caption{Bridging the Image History to the Image Response.}
	\label{fig:method3}
\vspace{-35pt}
\end{wrapfigure}

\begin{figure*}[t]
	\centering
	\includegraphics[width=1.0\linewidth]{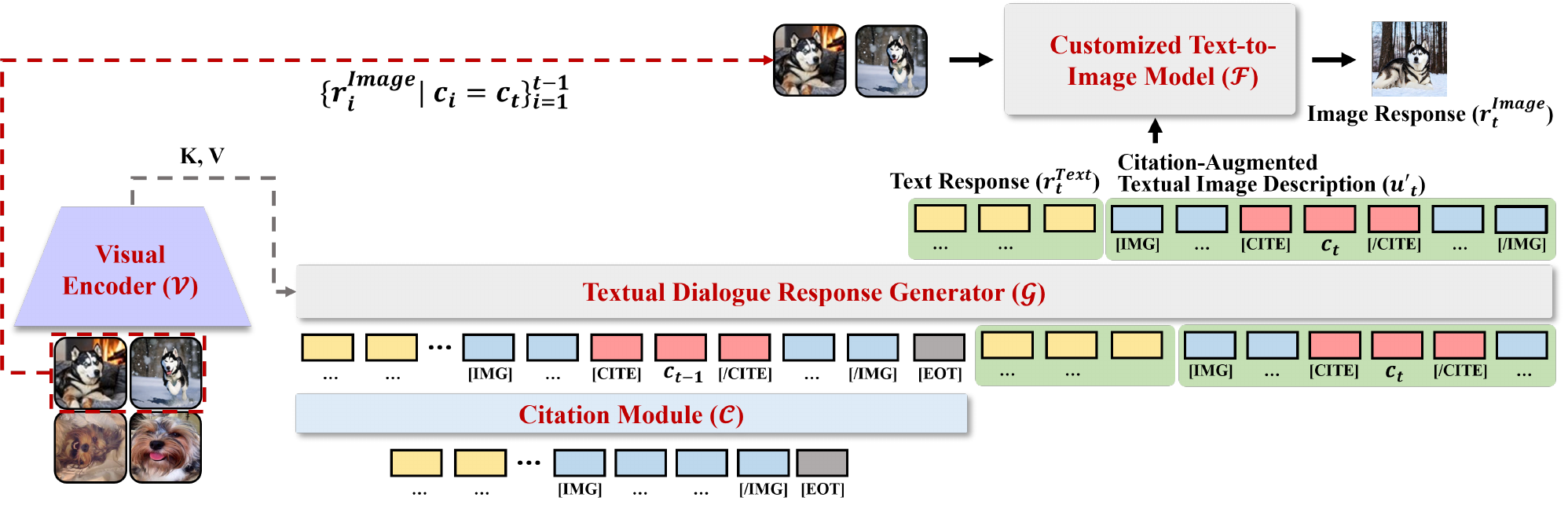}
	\caption{\textbf{Overall Inference Procedure.} During inference, the Textual Response Generator $\mathcal{G}$ generates the citation-augmented textual image description $u'_t$. By utilizing its citation tag $c_t$, object consistency can be maintained by feeding all preceding image responses with the identical citation tag into a Customized Text-to-Image Model $\mathcal{F}$.} 
	\label{fig:inference}
\end{figure*}

This section outlines our inference procedure, which employs Customized Text-to-Image Model \cite{blipdiffusion} in conjunction with citation-augmented textual image descriptions (Figure \ref{fig:inference}). This bridges the image history $r_{1:t-1}^{\text{Image}}$ and image response $r_{t}^{\text{Image}}$ (as depicted in Figure \ref{fig:method3}) allowing for consistent generation of the image response. 

\subsubsection{Inference}
For an incoming dialogue context $D=\{(r_i^{\text{Text}},r_i^{\text{Image}})\}_{i=1}^{t-1}$, we initially construct the sequence of textual image descriptions and corresponding text responses as $\{r_1^{\text{Text}},u_1,\ldots,r_{t-1}^{\text{Text}},u_{t-1}\}$. Our Citation Module $\mathcal{C}$ then augments the textual image descriptions with citation tags to create $\{r_1^{\text{Text}},u'_1,\ldots,r_{t-1}^{\text{Text}},u'_{t-1}\}$, which serves as the input for the Textual Response Generator $\mathcal{G}$. During inference, if $\mathcal{G}$ predicts an \texttt{[IMG]} token, it initiates the generation of a citation-augmented textual image description $u'_t$. We extract the citation $c_t$ and the core image description $u_t$ from $u'_t$. Subsequently, the image description $u_t$ is input to our Customized Text-to-Image Model $\mathcal{F}$, along with all preceding image responses sharing the same citation $c_t$ as follows: 
\begin{equation}
  r_t^{\text{Image}} = \mathcal{F}\left(u_t \mid \{r_i^{\text{Image}} \mid c_i = c_t\}_{i=1}^{t-1}\right).
  \label{eq:image_response}
\end{equation}
This approach ensures the consistent generation of specific objects across the conversation.
\section{Experiments}
This section presents the implementation details, evaluation benchmarks, and the experimental results of our approach.
Section \ref{sec4.1} outlines the implementation specifics. In Section \ref{sec4.2}, we assess the overall quality of our model against the standard benchmarks established in previous works \cite{related_photochat ,mmdialog}. Section \ref{sec4.3} is dedicated to evaluating the image grounding effectiveness of our model. Finally, Section \ref{sec4.4} examines the consistency of the image responses generated by our system. 

\subsection{Experimental Setup}
\label{sec4.1}
\subsubsection{Dataset} As mentioned in Section \ref{related_works}, our benchmark datasets include ImageChat \cite{imagechat}, PhotoChat \cite{related_photochat}, and MMDialog \cite{mmdialog}. Specifically, for the overall multimodal dialogue evaluation in Section \ref{sec4.2}, we train and evaluate using the PhotoChat and MMDialog. For the image grounding evaluation in Section \ref{sec4.3}, we use ImageChat. For the image consistency evaluation in Section \ref{sec4.4}, we create the MDIC dataset by hand-labeling a subset of MMDialog test set. For Section \ref{sec4.3} and Section \ref{sec4.4}, we use the model trained on MMDialog from Section \ref{sec4.2}. 

\subsubsection{Implementation Details} For the Textual Dialogue Response Generator $\mathcal{G}$ and the Visual Encoder $\mathcal{V}$, we initialize with the pre-trained OpenFlamingo 4B model\footnote{\href{https://huggingface.co/openflamingo/OpenFlamingo-4B-vitl-rpj3b-langinstruct}{https://huggingface.co/openflamingo/OpenFlamingo-4B-vitl-rpj3b-langinstruct}} \cite{openflamingo}. During the fine-tuning phase, we employ special tokens to structure our inputs and outputs: \texttt{[IMG]} and \texttt{[/IMG]} encapsulate textual image descriptions, while \texttt{[EOT]} signifies the end of conversation turns. The BLIP2-flan-t5-xl model \cite{blip2} is used for converting image responses to corresponding textual image descriptions. Additionally, our Citation Module $\mathcal{C}$ uses \texttt{[CITE]} and \texttt{[/CITE]} tokens to mark the beginning and end of citations linked to key objects within the textual image descriptions.

As noted in Section~\ref{3.2}, the Citation Module $\mathcal{C}$ is composed of four key components: the POS tagging processor (\(\mathcal{P}\)), the open-set object detector (\(\mathcal{M}\)), the segmentation model (\(\mathcal{S}\)), and the visual feature extractor (\(\mathcal{E}\)). Specifically, we employ spaCy~\cite{spacy2} for \(\mathcal{P}\), GroundingDino~\cite{dino} for \(\mathcal{M}\), the Segment Anything Model (SAM)~\cite{sam} for \(\mathcal{S}\), and DINOv2~\cite{dinov2} for \(\mathcal{E}\). The similarity threshold $\tau$ is set to 0.6. No further learning is done for these modules; they are utilized as pre-trained components within our system for citation tagging. For our customized text-to-image generation model, $\mathcal{F}$, we used BLIP-Diffusion \cite{blipdiffusion} when conditioning on the input image and the standard Stable Diffusion 2.1 \cite{stablediffusion} when there is no input image conditioning. 

\subsubsection{Learning Details}
Let us denote $\{\theta_\mathcal{V}, \theta_{\mathcal{G}_v}, \theta_{\mathcal{G}_l}\}$ as the parameters of the perceiver resampler of the Visual Encoder $\mathcal{V}$, the visual cross-attention layers of the Textual Dialogue Response Generator $\mathcal{G}$, and the language model layers of $\mathcal{G}$, respectively.

In the first stage of training, we train $\theta_{\mathcal{G}_l}$. The batch size is set to 256 with a maximum token length set to 256. In the second stage of training, we jointly train $\theta_\mathcal{V}$ and $\theta_{\mathcal{G}_v}$. The batch size is set to 128 with a maximum token length set to 512. Both the first and second stage is trained by minimizing the next token prediction loss (Eq. \ref{eq:training}) using the AdamW optimizer \cite{adamw} with a learning rate set to 1e-4. The trainings were conducted using 16 x NVIDIA A100 80GB PCIe.

\subsection{Multimodal Dialogue Evaluation}
\label{sec4.2}
\subsubsection{Evaluation Dataset} We evaluate the overall performance of our BI-MDRG system on the test set of PhotoChat \cite{related_photochat} and MMDialog \cite{mmdialog} dataset. PhotoChat contains a single image per dialogue, while the MMDialog includes dialogues with multiple images across turns, offering a more complex context for assessing our system's capability. We perform the tests for all turns except the first turn of each dialogue and consider all previous turns as context.

\subsubsection{Evaluation Metric} The performance evaluation is carried out using automatic metrics across four key dimensions: (1) Image Intent Prediction - assessing the need for an image response in the current turn; (2) Image Response Quality; (3) Textual Image Description Quality; and (4) Text Response Generation. We employ the F1 metric for Image Intent Prediction, following the binary classification approach detailed in \cite{divter, mmdialog}. BLEU \cite{bleu} and ROUGE \cite{ROUGE} metrics are used for evaluating both the Textual Image Descriptions and Text Responses. The Image Response Quality is measured using the Inception Score (IS) \cite{is_score}, in line with \cite{mmdialog}.

\subsubsection{Baselines} The BI-MDRG system is compared against Divter \cite{divter}, a prior Multimodal Dialogue Response Generator that utilizes DialoGPT (762M) \cite{dialogGPT} as its language model backbone. For a fair comparison, we also train Divter on the same backbone Language Model as our BI-MDRG, denoted as Divter$_\text{LLM}$ (3B). Moreover, for the MMDialog dataset, we compare with MiniGPT5 \cite{minigpt5} which uses a 9B backbone VLM. For Intent-prediction baselines, we also include BERT-base, T5-3B reported in \cite{related_photochat} for the PhotoChat dataset and PaCE reported in \cite{pace} for the MMDialog dataset.

\begin{table}[t]
\scriptsize
  \caption{Automatic evaluation results of BI-MDRG on the PhotoChat and the MMDialog test set. Numbers in bold represent the best scores.}
  \centering
  \begin{tabular}{l || c c c c m{0.8cm} m{0.8cm} m{0.5cm} c c c c c c}
    \toprule
            \multirow{2}{*}{Model} & Intent &  &\multicolumn{1}{c}{Image Response} &  &\multicolumn{4}{c}{Textual Image Description} &  &\multicolumn{4}{c}{Text Response}\\
		  & F1 &  & IS  &  & \hfil$\text{B1}$ & \hfil$\text{B2}$ & \hfil$\text{R-1}$ & \hfil$\text{R-L}$ &  & B1 & B2 & R1 & R-L\\
    \midrule

\rowcolor{gray!30}\multicolumn{14}{c}{\textbf{PhotoChat Dataset \cite{related_photochat}}} \\
\midrule
    \midrule
    BERT-base \cite{related_photochat} & 53.2 &  & - &  & \hfil$\text{-}$ & \hfil$\text{-}$ & \hfil$\text{-}$ & \hfil$\text{-}$ &  & - & - & - & -\\
    T5-3B \cite{related_photochat} & \textbf{58.9} &  & - &  & \hfil$\text{-}$ & \hfil$\text{-}$ & \hfil$\text{-}$ & \hfil$\text{-}$ &  & - & - & - & -\\
    Divter \cite{divter} & 56.2 &  & 15.8 &  & \hfil$\text{15.1}$ & \hfil$\text{11.4}$ & \hfil$\text{-}$ & 15.8 &  & 6.52 & 1.66 & - & 5.69\\
    Divter$_{\text{LLM}}$ & 54.1 &  & 16.1   &  & \hfil$\text{41.3}$ & \hfil$\text{27.1}$ & 43.3 & 41.6 &  & 11.4 & 4.75 & 11.2 & 10.8\\
     \textbf{BI-MDRG} & 55.7 &  & \textbf{16.7}  &   & \hfil$\text{\textbf{42.1}}$ & \hfil$\text{\textbf{28.2}}$ & \textbf{44.6} & \textbf{42.5} &  & \textbf{12.4} & \textbf{5.12} & \textbf{12.1} & \textbf{11.2}\\
\midrule

\rowcolor{gray!30}\multicolumn{14}{c}{\textbf{MMDialog Dataset \cite{mmdialog}}} \\
\midrule
\midrule
    PaCE \cite{pace} & \textbf{77.6} &  & \hfil$\text{-}$ &  & \hfil$\text{-}$ & \hfil$\text{-}$ & \hfil$\text{-}$& - &  & \hfil$\text{-}$ & \hfil$\text{-}$ & - & \hfil$\text{-}$\\
    Divter \cite{mmdialog} & 71.8 &  & 20.5 &  & \hfil$\text{-}$ & \hfil$\text{-}$ & \hfil$\text{-}$& - &  & 9.44 & 7.45 & - & 11.2\\
     MiniGPT5 \cite{minigpt5} & - &  & 20.2 &  & \hfil$\text{-}$ & \hfil$\text{-}$ & \hfil$\text{-}$& - &  & \textbf{29.1} & 19.5 & - & 12.1\\
    Divter$_{\text{LLM}}$ & 67.3 &  & 21.0   &  & \hfil$\text{44.2}$ & \hfil$\text{35.7}$ & 45.5 & 43.6 &  & 21.3 & 16.2 & 20.4 & 19.4\\
     \textbf{BI-MDRG} & 70.5 &  & \textbf{22.4}  &   & \hfil$\text{\textbf{52.2}}$ & \hfil$\text{\textbf{44.7}}$ & \textbf{53.2} & \textbf{51.6} &  & 27.6 & \textbf{23.5} & \textbf{25.7} & \textbf{24.8}\\   
    \bottomrule
  \end{tabular}

  \label{tab:exp_main_result}
\end{table}

\subsubsection{Results}
Table \ref{tab:exp_main_result} presents the evaluation results, where our BI-MDRG system demonstrates outstanding performance on both datasets. In Textual Image Description and Text Response Generation, BI-MDRG achieves the highest BLEU and ROUGE metrics scores in most cases, indicating the system's proficiency in generating relevant and coherent image responses and text responses. Notably, BI-MDRG, which uses a 4B backbone VLM, outperforms MiniGPT5, which utilizes a 9B backbone VLM.

\subsection{Image Grounding Evaluation}
\label{sec4.3}
\subsubsection{Evaluation Dataset} To evaluate the ability to generate image-grounded textual responses, we utilize the ImageChat dataset \cite{imagechat}, consisting of dialogues centered around a single image (an example is shown in Appendix C.4). Since the conversations are grounded in an image, the dataset is suitable for the evaluation of the image grounding capability of the model. We use the same model trained on the MMDialog dataset in Section \ref{sec4.2} without further tuning on the ImageChat train set.

\subsubsection{Evaluation Metric} 
In order to evaluate the image grounding capability, we use BLEU and ROUGE as the metrics. We perform the evaluation only on the last turn of each dialogue and consider all previous turns as the input context.
\begin{wraptable}{r}{0.45\textwidth}
\small
  \caption{Automatic evaluation results on the ImageChat test set. Numbers in bold represent the best scores.\\}
  \centering
  \begin{tabular}{l || c c c}
    \toprule
    Model & B1 & R-1 & R-L \\
\midrule
\rowcolor{gray!30}\multicolumn{4}{c}{\textbf{ImageChat Dataset \cite{imagechat}}} \\
    \midrule
    \midrule
    Divter$_{\text{LLM}}$ & 8.6 & 10.3 & 9.6 \\
    BI-MDRG$_{\text{w/o mask}}$ & 10.0 & 11.1 & 10.2 \\
    \textbf{BI-MDRG} & \textbf{10.9} & \textbf{11.7} & \textbf{10.9 }\\
    \bottomrule
  \end{tabular}
  \label{tab:imagechat}
\vspace{-30pt}
\end{wraptable}
\subsubsection{Results}
The performance of our BI-MDRG system on the ImageChat benchmark is summarized in Table \ref{tab:imagechat}. These results demonstrate the system's superior image grounding capabilities in text responses. BI-MDRG achieves the highest scores in BLEU-1, ROUGE-1, and ROUGE-L metrics, indicating its effectiveness in generating contextually relevant and coherent text responses grounded in the visual elements of the conversation. In comparison to Divter$_{\text{LLM}}$ and BI-MDRG$_{\text{w/o mask}}$, our BI-MDRG model shows an improvement, highlighting the significance of our proposed architectural enhancements and the multimodal causal attention mask in understanding and integrating visual context into text responses effectively. 

\begin{figure}[t]
	\centering
	\includegraphics[width=0.9\linewidth]{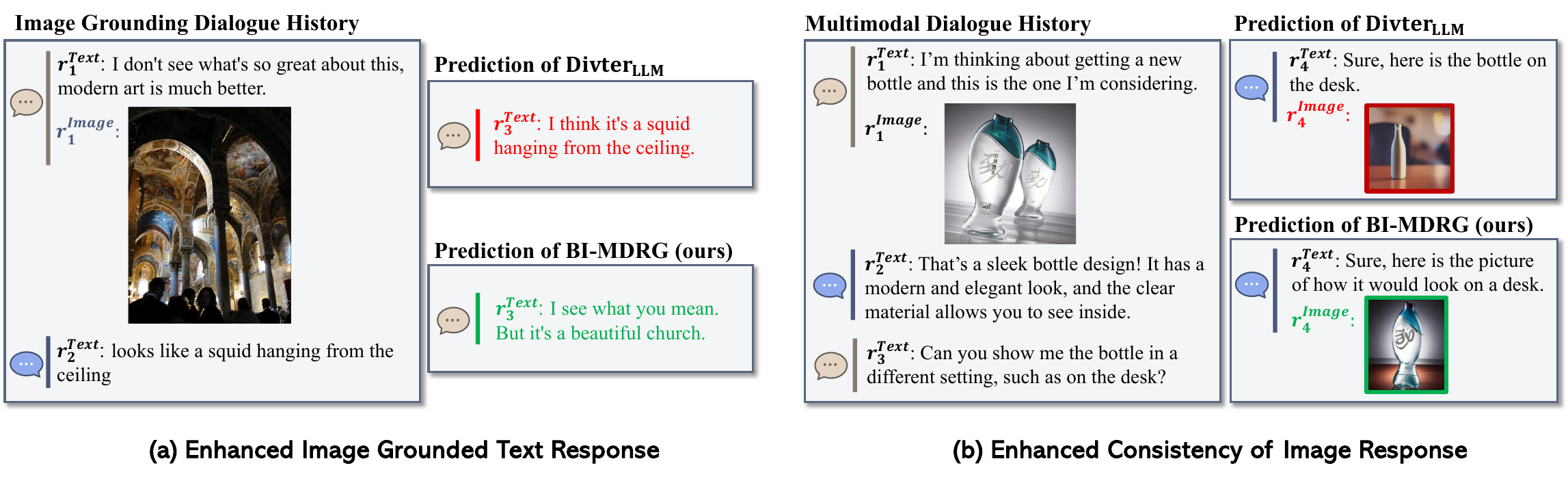}
    \caption{Example of predictions from Divter$_{\text{LLM}}$ \cite{divter} vs. BI-MDRG (ours). Additional examples can be found in Appendix E.}
	\label{fig:main_example}

\end{figure}

Moreover, we show an example of the model prediction between the Divter$_{\text{LLM}}$ and our BI-MDRG in Figure \ref{fig:main_example}-(a). Figure \ref{fig:main_example}-(a) (\textit{right-top}) shows that Divter$_{\mathbf{LLM}}$ produces contextually irrelevant text response due to its lack of image understanding. However, Figure \ref{fig:main_example}-(a) (\textit{right-bottom}) shows that since our BI-MDRG has access to the image, it can produce a text response grounded on the image.

\begin{figure*}[t!]
	\centering
	\includegraphics[width=1.0\linewidth]{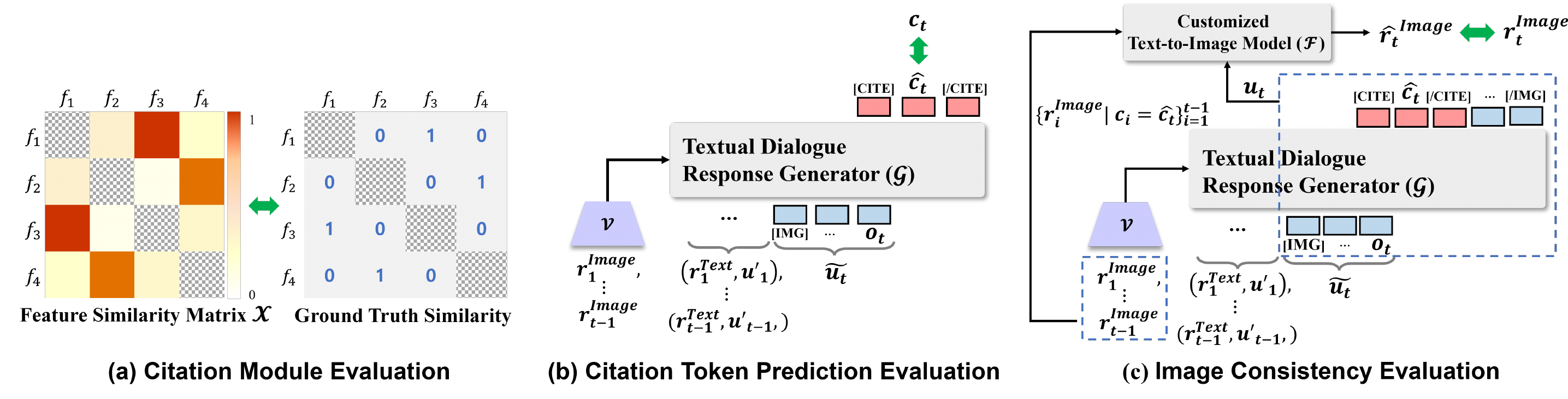}
	\caption{\textbf{Three evaluation aspects for image response consistency.} (a) We assess the accuracy of the pseudo-labels of citation tags created by the Citation Module. (b) We evaluate how well the model learns from these pseudo-labels and predict the citation tag $\hat{c_t}$, given the dialogue history, current text response $r_t^{\text{Text}}$, and $\tilde{u_t}$ which denotes the textual image description leading up to the primary object word $o_t$. (c) We evaluate the consistency of the resulting image response $\hat{r}_t^{\text{Image}}$.}
	\label{fig:evaluation}
\end{figure*}

\subsection{Image Response Consistency Evaluation}
\label{sec4.4}
We integrated the Citation Module into our BI-MDRG system to improve consistency in image responses within the dialogue. This module assigns citation tags to primary objects in textual image descriptions, tracking their presence across the dialogue. 
As shown in Figure \ref{fig:evaluation}, we focus on three aspects for evaluating the image response consistency: \textit{(a) Citation Module Evaluation}, \textit{(b) Citation Token Prediction Evaluation}, and \textit{(c) Image Consistency Evaluation}.
\subsubsection{Multimodal Dialogue Image Consistency (MDIC) Dataset Creation}
Due to the absence of benchmark datasets for evaluating the image consistency in multimodal dialogue scenarios, we created a dataset of 300 dialogues, each manually annotated to assign correct citation tags to objects based on their appearances in image responses. We use this created dataset for the subsequent evaluations. Details of the annotation process can be found in Appendix C.5.

\subsubsection{Citation Module Evaluation} 
Figure \ref{fig:evaluation}-(a) shows the evaluation method for the Citation Module. As described in Section \ref{3.2}, the module \(\mathcal{C}\) processes visual features \(\{f_1, \ldots, f_n\}\) corresponding to primary objects \(\{o_1, \ldots, o_n\}\) in image descriptions. The module generates citations based on the cosine similarity between these features. We constructed a similarity matrix \(\mathcal{X}\), where each element \(\mathcal{X}_{ij}\) represents the cosine similarity between the features \(f_i\) and \(f_j\). A feature pair \((f_i, f_j)\) is deemed `similar' if \(\mathcal{X}_{ij} \geq \tau\) and `not similar' otherwise, where \(\tau\) is a predetermined threshold. The module's performance was quantified using the F1 score, and we achieved an F1 score of 0.72, demonstrating the module's high accuracy in assigning pseudo-labels of citation tags for training.
\begin{wraptable}{r}{0.45\textwidth}
\scriptsize
  \caption{Citation Token Prediction (Acc.) and Image Consistency Evaluation (DINOv2). Numbers in bold represent the best scores.\\}
  \centering
  \begin{tabular}{l | c | c }
    \toprule
   Model & Acc. & DINOv2($\uparrow$) \\
    \midrule
    \rowcolor{gray!30}\multicolumn{3}{c}{\textbf{MDIC Dataset}} \\
    \midrule
    \midrule
    Divter$_{\text{LLM}}$ + LLMCite & 33.5 & 0.32\\
    \textbf{BI-MDRG (ours)} & \textbf{84.0} & \textbf{0.53} \\
    \bottomrule
  \end{tabular}
  \label{tab:citation_token_eval}
\vspace{-20pt}
\end{wraptable}
\subsubsection{Citation Token Prediction Evaluation} Our model, trained with citation-augmented textual image descriptions, can generate these descriptions with accurate citation tags for the primary objects in the descriptions. 
Figure \ref{fig:evaluation}-(b) shows the evaluation method where we assess our model's ability to predict the citation tags correctly. We input the ground truth sequence $\{r_1^{\text{Text}},u'_1, \ldots, r_{t}^{\text{Text}},\tilde{u_t}\}$ from the MDIC dataset along with the image response history $\{r_1^{\text{Image}}, \ldots, r_{t-1}^{\text{Image}}\}$. Here, $\tilde{u_t}$ denotes the textual image description leading up to the principal object word $o_t$. The goal is to verify whether the model's predicted citation $\hat{c_t}$ aligns with the actual citation $c_t$. The accuracy (Acc.) of our citation token prediction is presented in Table \ref{tab:citation_token_eval}. We compare with a baseline denoted as Divter$_{\text{LLM}}$ + LLMCite where we assign citation tags by instructing an external LLM (Details in Appendix D). 
The results show that BI-MDRG achieves 84.0\% accuracy compared to 33.5\% of the LLMCite baseline, showing the effectiveness of our model in identifying objects needing consistency.

\subsubsection{Image Consistency Evaluation} 
Figure \ref{fig:evaluation}-(c) shows the evaluation method for the Image Consistency Evaluation. To assess the consistency of the image response, we employ metrics commonly found in the literature on custom object text-to-image generation \cite{custom_diffusion,customize1,customize2,customize3,customize4}. 
DINO features outperform CLIP in capturing image alignment, specifically in distinguishing different objects of the same class \cite{customize4}. Therefore, for our evaluation, we measure image alignment scores using the DINOv2 \cite{dinov2} features. 

In our proposed framework, correctly using the citation tag is essential to ensure image consistency in the conversation. This involves correctly predicting a new citation tag or referencing an existing tag in history. Therefore, our evaluation includes not only scenarios where the model's predicted citation tag aligns with the ground truth (\(c_t = \hat{c_t}\)) and \(c_t\) exists in previous dialogue history, but also instances where the model's predicted citation tag do not match the ground truth (\(c_t \neq \hat{c_t}\)) and either \(\hat{c_t}\) or \(c_t\) is in previous dialogue history.

We report the DINOv2 score as an average of both the successful and failed citation cases mentioned above, where successful prediction positively influences the score, and failed prediction reduces it, offering a comprehensive measure of the model's performance in image consistency. Table \ref{tab:citation_token_eval} shows that our method successfully maintains the consistency of the image response in a dialogue setting, with the DINOv2 score achieving 0.53 compared to 0.32 of baseline. Furthermore, in Figure \ref{fig:main_example}-(b), we present a comparative example of enhanced image consistency of prediction from \textbf{BI-MDRG} compared to \textbf{Divter}$_{\mathbf{LLM}}$.

\section{Importance of Citation Tags for Image Consistency}
\label{section5}

Due to powerful pre-trained models for text-to-text and text-to-image, adopting text as an intermediary for image responses is a practical solution for the MDRG \cite{divter} task. However, due to the inherent information loss of images during this process, achieving \textit{image consistency is infeasible without a targeted framework for consistency maintenance.}

Table \ref{tab:1} shows image consistency and dialogue response performance across various settings. Not using citations shows similar dialogue response performance (Intent, TID, TR) compared to LLMCite and our Citation Module. However, LLMCite improves image consistency (DINOv2) from 0.25 to 0.34 (4B) and 0.26 to 0.33 (9B). Our Citation Module further boosts this from 0.34 to 0.53, indicating its importance for image consistency without affecting dialogue response performance. Notably, scaling the model size (4B to 9B) improves textual response but fails to maintain image consistency without our citation framework, as also evident by ChatGPT’s shortcomings in Appendix E. 

\begin{table}[t]
  \caption{\textbf{Evaluation of the citation framework on various settings.} The image consistency is measured using DINOv2 ($\uparrow$) on MDIC dataset, and TID (Textual Image Description) and TR (Text Response) are measured on MMDialog \cite{mmdialog} test set. Custom uses customized text-to-image generation with image conditioning, while Diffusion employs a standard text-to-image diffusion model without conditioning. The citation framework allows to selectively use between these for consistent image response. \textit{The first row represents the performance of BI-MDRG (ours).}}
\vspace{-5pt}
  \centering
  \scriptsize
  \setlength{\tabcolsep}{1.2pt}
  \begin{tabular}{c c c || c | c c  c c c c c c c c}
  \toprule

              \multirow{2}{*}{Citation} & VLM &\multirow{2}{*}{Diffusion} & \multirow{2}{*}{DINOv2} & & \multicolumn{4}{c}{TID}  & & \multicolumn{4}{c}{TR} \\
		  & Size &  &  & & B1 & B2 & R-1 & R-L & & B1 & B2 & R-1 & R-L \\
   \midrule

    \textbf{Citation Module} & 4B & Custom+Diffusion &  \cellcolor{green!30}\textbf{0.53} & & 52.2 & 44.7 & 53.2 & 51.6 & & 27.6 & 23.5 & 25.7 & 24.8\\
    \midrule
    \midrule
      LLMCite & 4B & Custom+Diffusion & \cellcolor{green!30}0.34 & & 52.0 & 44.9 & 53.3 & 51.3 & & 27.4 & 23.4 & 25.8 & 24.6\\
     x & 4B & Diffusion & \cellcolor{green!30}0.25 & & 52.0 & 44.9 & 53.3 & 51.3 & & 27.4 & 23.4 & 25.8 & 24.6\\
    \midrule
     LLMCite & 9B & Custom+Diffusion & \cellcolor{green!30}0.33 & & \textbf{56.0} & \textbf{48.7} & \textbf{54.5} & \textbf{52.1} & & \textbf{30.3} & \textbf{25.7} & \textbf{28.8} & \textbf{26.8}\\
     x & 9B & Diffusion & \cellcolor{green!30}0.26 & & \textbf{56.0} & \textbf{48.7} & \textbf{54.5} & \textbf{52.1} & & \textbf{30.3} & \textbf{25.7} & \textbf{28.8} & \textbf{26.8}\\
    \bottomrule
  \end{tabular}
  \label{tab:1}
  \vspace{-10pt}
\end{table}
\section{Conclusion}
This paper presents BI-MDRG, a novel framework for Multimodal Dialogue Response Generation (MDRG) aimed at bridging the image history for enhanced text and image response. Our model's innovative use of image history to inform both text and image responses addresses fundamental limitations in previous methodologies, particularly in maintaining consistency in multimodal interactions. 
The effectiveness of BI-MDRG has been demonstrated through rigorous evaluations using multiple benchmark datasets and a custom-annotated dataset.

\section*{Acknowledgement} This work was supported by a grant of the KAIST-KT joint research project through AI2X Lab., Tech Innovation Group, funded by KT (No. D23000019, Developing Visual and Language Capabilities for AI-Based Dialogue Systems), and by Institute for Information \& communications Technology Planning \& Evaluation (IITP) grant funded by the Korea government(MSIT) (No. 2021-0-01381, Development of Causal AI through Video Understanding and Reinforcement Learning, and Its Applications to Real Environments).

%
%
\bibliographystyle{splncs04}
\bibliography{main}

\begin{thebibliography}{10}
\providecommand{\url}[1]{\texttt{#1}}
\providecommand{\urlprefix}{URL }
\providecommand{\doi}[1]{https://doi.org/#1}

\bibitem{related_avsd}
Alamri, H., Cartillier, V., Das, A., Wang, J., Cherian, A., Essa, I., Batra, D., Marks, T.K., Hori, C., Anderson, P., Lee, S., Parikh, D.: Audio-visual scene-aware dialog. In: Proceedings of the IEEE Conference on Computer Vision and Pattern Recognition (2019)

\bibitem{flamingo}
Alayrac, J.B., Donahue, J., Luc, P., Miech, A., Barr, I., Hasson, Y., Lenc, K., Mensch, A., Millican, K., Reynolds, M., Ring, R., Rutherford, E., Cabi, S., Han, T., Gong, Z., Samangooei, S., Monteiro, M., Menick, J., Borgeaud, S., Brock, A., Nematzadeh, A., Sharifzadeh, S., Binkowski, M., Barreira, R., Vinyals, O., Zisserman, A., Simonyan, K.: Flamingo: a visual language model for few-shot learning. In: Oh, A.H., Agarwal, A., Belgrave, D., Cho, K. (eds.) Advances in Neural Information Processing Systems (2022), \url{https://openreview.net/forum?id=EbMuimAbPbs}

\bibitem{retrieval_2}
Anonymous: Chatsearch: a dataset and a generative retrieval model for general conversational image retrieval (2023), \url{https://openreview.net/forum?id=0unbjYPmbC}

\bibitem{openflamingo}
Awadalla, A., Gao, I., Gardner, J., Hessel, J., Hanafy, Y., Zhu, W., Marathe, K., Bitton, Y., Gadre, S., Sagawa, S., Jitsev, J., Kornblith, S., Koh, P.W., Ilharco, G., Wortsman, M., Schmidt, L.: Openflamingo: An open-source framework for training large autoregressive vision-language models. arXiv preprint arXiv:2308.01390  (2023)

\bibitem{diffusion1}
Balaji, Y., Nah, S., Huang, X., Vahdat, A., Song, J., Zhang, Q., Kreis, K., Aittala, M., Aila, T., Laine, S., Catanzaro, B., Karras, T., Liu, M.Y.: ediff-i: Text-to-image diffusion models with ensemble of expert denoisers. arXiv preprint arXiv:2211.01324  (2022)

\bibitem{chen2023anydoor}
Chen, X., Huang, L., Liu, Y., Shen, Y., Zhao, D., Zhao, H.: Anydoor: Zero-shot object-level image customization. arXiv preprint arXiv:2307.09481  (2023)

\bibitem{customize1}
Cohen, N., Gal, R., Meirom, E.A., Chechik, G., Atzmon, Y.: “this is my unicorn, fluffy”: Personalizing frozen vision-language representations. In: European Conference on Computer Vision. pp. 558--577. Springer (2022)

\bibitem{related_visdialog}
Das, A., Kottur, S., Gupta, K., Singh, A., Yadav, D., Moura, J.M., Parikh, D., Batra, D.: Visual dialog. In: Proceedings of the IEEE conference on computer vision and pattern recognition. pp. 326--335 (2017)

\bibitem{mmdialog}
Feng, J., Sun, Q., Xu, C., Zhao, P., Yang, Y., Tao, C., Zhao, D., Lin, Q.: {MMD}ialog: A large-scale multi-turn dialogue dataset towards multi-modal open-domain conversation. In: Rogers, A., Boyd-Graber, J., Okazaki, N. (eds.) Proceedings of the 61st Annual Meeting of the Association for Computational Linguistics (Volume 1: Long Papers). pp. 7348--7363. Association for Computational Linguistics, Toronto, Canada (Jul 2023). \doi{10.18653/v1/2023.acl-long.405}, \url{https://aclanthology.org/2023.acl-long.405}

\bibitem{customize2}
Gal, R., Alaluf, Y., Atzmon, Y., Patashnik, O., Bermano, A.H., Chechik, G., Cohen-Or, D.: An image is worth one word: Personalizing text-to-image generation using textual inversion (2022). \doi{10.48550/ARXIV.2208.01618}, \url{https://arxiv.org/abs/2208.01618}

\bibitem{related_modeling2}
Gan, Z., Cheng, Y., Kholy, A., Li, L., Liu, J., Gao, J.: Multi-step reasoning via recurrent dual attention for visual dialog. In: Korhonen, A., Traum, D., M{\`a}rquez, L. (eds.) Proceedings of the 57th Annual Meeting of the Association for Computational Linguistics. pp. 6463--6474. Association for Computational Linguistics, Florence, Italy (Jul 2019). \doi{10.18653/v1/P19-1648}, \url{https://aclanthology.org/P19-1648}

\bibitem{related_ytd}
Han, S., Hessel, J., Dziri, N., Choi, Y., Yu, Y.: Champagne: Learning real-world conversation from large-scale web videos. arXiv preprint arXiv:2303.09713  (2023)

\bibitem{spacy2}
Honnibal, M., Montani, I.: {spaCy 2}: Natural language understanding with {B}loom embeddings, convolutional neural networks and incremental parsing (2017), to appear

\bibitem{kim2021structuredcoreferencegraphattention}
Kim, J., Yoon, S., Kim, D., Yoo, C.D.: Structured co-reference graph attention for video-grounded dialogue (2021), \url{https://arxiv.org/abs/2103.13361}

\bibitem{sam}
Kirillov, A., Mintun, E., Ravi, N., Mao, H., Rolland, C., Gustafson, L., Xiao, T., Whitehead, S., Berg, A.C., Lo, W.Y., Doll{\'a}r, P., Girshick, R.: Segment anything. arXiv:2304.02643  (2023)

\bibitem{retrieval_3}
Koh, J.Y., Salakhutdinov, R., Fried, D.: Grounding language models to images for multimodal inputs and outputs. ICML  (2023)

\bibitem{koo2024wavelet}
Koo, G., Yoon, S., Yoo, C.D.: Wavelet-guided acceleration of text inversion in diffusion-based image editing. In: ICASSP 2024-2024 IEEE International Conference on Acoustics, Speech and Signal Processing (ICASSP). pp. 4380--4384. IEEE (2024)

\bibitem{customize3}
Kumari, N., Zhang, B., Zhang, R., Shechtman, E., Zhu, J.Y.: Multi-concept customization of text-to-image diffusion  (2023)

\bibitem{custom_diffusion}
Kumari, N., Zhang, B., Zhang, R., Shechtman, E., Zhu, J.Y.: Multi-concept customization of text-to-image diffusion (2023)

\bibitem{related_mmdd}
Lee, N., Shin, S., Choo, J., Choi, H.J., Myaeng, S.H.: Constructing multi-modal dialogue dataset by replacing text with semantically relevant images. In: Proceedings of the 59th Annual Meeting of the Association for Computational Linguistics and the 11th International Joint Conference on Natural Language Processing (Volume 2: Short Papers). pp. 897--906. Association for Computational Linguistics, Online (Aug 2021), \url{https://aclanthology.org/2021.acl-short.113}

\bibitem{related_dialogcc}
Lee, Y.J., Ko, B., Kim, H.G., Choi, H.J.: Dialogcc: Large-scale multi-modal dialogue dataset. arXiv preprint arXiv:2212.04119  (2022)

\bibitem{retrieval_1}
Levy, M., Ben-Ari, R., Darshan, N., Lischinski, D.: Chatting makes perfect: Chat-based image retrieval. In: Oh, A., Neumann, T., Globerson, A., Saenko, K., Hardt, M., Levine, S. (eds.) Advances in Neural Information Processing Systems. vol.~36, pp. 61437--61449. Curran Associates, Inc. (2023), \url{https://proceedings.neurips.cc/paper_files/paper/2023/file/c1b3d1e2cf53bb28cabd801bd58b3521-Paper-Conference.pdf}

\bibitem{blipdiffusion}
Li, D., Li, J., Hoi, S.C.H.: Blip-diffusion: Pre-trained subject representation for controllable text-to-image generation and editing (2023)

\bibitem{blip2}
Li, J., Li, D., Savarese, S., Hoi, S.: Blip-2: Bootstrapping language-image pre-training with frozen image encoders and large language models. arXiv preprint arXiv:2301.12597  (2023)

\bibitem{pace}
Li, Y., Hui, B., Yin, Z., Yang, M., Huang, F., Li, Y.: {P}a{CE}: Unified multi-modal dialogue pre-training with progressive and compositional experts. In: Rogers, A., Boyd-Graber, J., Okazaki, N. (eds.) Proceedings of the 61st Annual Meeting of the Association for Computational Linguistics (Volume 1: Long Papers). pp. 13402--13416. Association for Computational Linguistics, Toronto, Canada (Jul 2023). \doi{10.18653/v1/2023.acl-long.749}, \url{https://aclanthology.org/2023.acl-long.749}

\bibitem{ROUGE}
Lin, C.Y.: {ROUGE}: A package for automatic evaluation of summaries. In: Text Summarization Branches Out. pp. 74--81. Association for Computational Linguistics, Barcelona, Spain (Jul 2004), \url{https://aclanthology.org/W04-1013}

\bibitem{related_tiktalk}
Lin, H., Ruan, L., Xia, W., Liu, P., Wen, J., Xu, Y., Hu, D., Song, R., Zhao, W.X., Jin, Q., et~al.: Tiktalk: A video-based dialogue dataset for multi-modal chitchat in real world. In: Proceedings of the 31st ACM International Conference on Multimedia. pp. 1303--1313 (2023)

\bibitem{dino}
Liu, S., Zeng, Z., Ren, T., Li, F., Zhang, H., Yang, J., Li, C., Yang, J., Su, H., Zhu, J., et~al.: Grounding dino: Marrying dino with grounded pre-training for open-set object detection. arXiv preprint arXiv:2303.05499  (2023)

\bibitem{adamw}
Loshchilov, I., Hutter, F.: Decoupled weight decay regularization. In: International Conference on Learning Representations

\bibitem{ma2023subject}
Ma, J., Liang, J., Chen, C., Lu, H.: Subject-diffusion: Open domain personalized text-to-image generation without test-time fine-tuning. arXiv preprint arXiv:2307.11410  (2023)

\bibitem{related_openvidial}
Meng, Y., Wang, S., Han, Q., Sun, X., Wu, F., Yan, R., Li, J.: Openvidial: A large-scale, open-domain dialogue dataset with visual contexts. arXiv preprint arXiv:2012.15015  (2020)

\bibitem{related_modeling1}
Niu, Y., Zhang, H., Zhang, M., Zhang, J., Lu, Z., Wen, J.R.: Recursive visual attention in visual dialog. In: Proceedings of the IEEE/CVF Conference on Computer Vision and Pattern Recognition. pp. 6679--6688 (2019)

\bibitem{dinov2}
Oquab, M., Darcet, T., Moutakanni, T., Vo, H., Szafraniec, M., Khalidov, V., Fernandez, P., Haziza, D., Massa, F., El-Nouby, A., Assran, M., Ballas, N., Galuba, W., Howes, R., Huang, P.Y., Li, S.W., Misra, I., Rabbat, M., Sharma, V., Synnaeve, G., Xu, H., Jegou, H., Mairal, J., Labatut, P., Joulin, A., Bojanowski, P.: Dinov2: Learning robust visual features without supervision (2023)

\bibitem{bleu}
Papineni, K., Roukos, S., Ward, T., Zhu, W.J.: Bleu: A method for automatic evaluation of machine translation. In: Proceedings of the 40th Annual Meeting on Association for Computational Linguistics. p. 311–318. ACL '02, Association for Computational Linguistics, USA (2002). \doi{10.3115/1073083.1073135}, \url{https://doi.org/10.3115/1073083.1073135}

\bibitem{related_meld}
Poria, S., Hazarika, D., Majumder, N., Naik, G., Cambria, E., Mihalcea, R.: {MELD}: A multimodal multi-party dataset for emotion recognition in conversations. In: Korhonen, A., Traum, D., M{\`a}rquez, L. (eds.) Proceedings of the 57th Annual Meeting of the Association for Computational Linguistics. pp. 527--536. Association for Computational Linguistics, Florence, Italy (Jul 2019). \doi{10.18653/v1/P19-1050}, \url{https://aclanthology.org/P19-1050}

\bibitem{related_modeling3}
Qi, J., Niu, Y., Huang, J., Zhang, H.: Two causal principles for improving visual dialog. In: Proceedings of the IEEE/CVF conference on computer vision and pattern recognition. pp. 10860--10869 (2020)

\bibitem{gpt2}
Radford, A., Wu, J., Child, R., Luan, D., Amodei, D., Sutskever, I.: Language models are unsupervised multitask learners  (2019)

\bibitem{stablediffusion}
Rombach, R., Blattmann, A., Lorenz, D., Esser, P., Ommer, B.: High-resolution image synthesis with latent diffusion models. In: Proceedings of the IEEE/CVF Conference on Computer Vision and Pattern Recognition (CVPR). pp. 10684--10695 (June 2022)

\bibitem{customize4}
Ruiz, N., Li, Y., Jampani, V., Pritch, Y., Rubinstein, M., Aberman, K.: Dreambooth: Fine tuning text-to-image diffusion models for subject-driven generation. In: Proceedings of the IEEE/CVF Conference on Computer Vision and Pattern Recognition. pp. 22500--22510 (2023)

\bibitem{diffusion2}
Saharia, C., Chan, W., Saxena, S., Li, L., Whang, J., Denton, E., Ghasemipour, S.K.S., Gontijo-Lopes, R., Ayan, B.K., Salimans, T., Ho, J., Fleet, D.J., Norouzi, M.: Photorealistic text-to-image diffusion models with deep language understanding. In: Oh, A.H., Agarwal, A., Belgrave, D., Cho, K. (eds.) Advances in Neural Information Processing Systems (2022), \url{https://openreview.net/forum?id=08Yk-n5l2Al}

\bibitem{is_score}
Salimans, T., Goodfellow, I., Zaremba, W., Cheung, V., Radford, A., Chen, X., Chen, X.: Improved techniques for training gans. In: Lee, D., Sugiyama, M., Luxburg, U., Guyon, I., Garnett, R. (eds.) Advances in Neural Information Processing Systems. vol.~29. Curran Associates, Inc. (2016), \url{https://proceedings.neurips.cc/paper_files/paper/2016/file/8a3363abe792db2d8761d6403605aeb7-Paper.pdf}

\bibitem{imagechat}
Shuster, K., Humeau, S., Bordes, A., Weston, J.: Image-chat: Engaging grounded conversations. In: Jurafsky, D., Chai, J., Schluter, N., Tetreault, J. (eds.) Proceedings of the 58th Annual Meeting of the Association for Computational Linguistics. pp. 2414--2429. Association for Computational Linguistics, Online (Jul 2020). \doi{10.18653/v1/2020.acl-main.219}, \url{https://aclanthology.org/2020.acl-main.219}

\bibitem{divter}
Sun, Q., Wang, Y., Xu, C., Zheng, K., Yang, Y., Hu, H., Xu, F., Zhang, J., Geng, X., Jiang, D.: Multimodal dialogue response generation. In: Proceedings of the 60th Annual Meeting of the Association for Computational Linguistics (Volume 1: Long Papers). pp. 2854--2866. Association for Computational Linguistics, Dublin, Ireland (May 2022). \doi{10.18653/v1/2022.acl-long.204}, \url{https://aclanthology.org/2022.acl-long.204}

\bibitem{related_openvidial2}
Wang, S., Meng, Y., Li, X., Sun, X., Ouyang, R., Li, J.: Openvidial 2.0: A larger-scale, open-domain dialogue generation dataset with visual contexts. arXiv preprint arXiv:2109.12761  (2021)

\bibitem{yoon2023hear}
Yoon, S., Kim, D., Yoon, E., Yoon, H.S., Kim, J., Yoo, C.D.: Hear: Hearing enhanced audio response for video-grounded dialogue. arXiv preprint arXiv:2312.09736  (2023)

\bibitem{yoon-etal-2022-information}
Yoon, S., Yoon, E., Yoon, H.S., Kim, J., Yoo, C.: Information-theoretic text hallucination reduction for video-grounded dialogue. In: Goldberg, Y., Kozareva, Z., Zhang, Y. (eds.) Proceedings of the 2022 Conference on Empirical Methods in Natural Language Processing. pp. 4182--4193. Association for Computational Linguistics, Abu Dhabi, United Arab Emirates (Dec 2022). \doi{10.18653/v1/2022.emnlp-main.280}, \url{https://aclanthology.org/2022.emnlp-main.280}

\bibitem{related_photochat}
Zang, X., Liu, L., Wang, M., Song, Y., Zhang, H., Chen, J.: {P}hoto{C}hat: A human-human dialogue dataset with photo sharing behavior for joint image-text modeling. In: Zong, C., Xia, F., Li, W., Navigli, R. (eds.) Proceedings of the 59th Annual Meeting of the Association for Computational Linguistics and the 11th International Joint Conference on Natural Language Processing (Volume 1: Long Papers). pp. 6142--6152. Association for Computational Linguistics, Online (Aug 2021). \doi{10.18653/v1/2021.acl-long.479}, \url{https://aclanthology.org/2021.acl-long.479}

\bibitem{dialogGPT}
Zhang, Y., Sun, S., Galley, M., Chen, Y.C., Brockett, C., Gao, X., Gao, J., Liu, J., Dolan, B.: Dialogpt: Large-scale generative pre-training for conversational response generation. In: ACL, system demonstration (2020)

\bibitem{related_m3ed}
Zhao, J., Zhang, T., Hu, J., Liu, Y., Jin, Q., Wang, X., Li, H.: {M}3{ED}: Multi-modal multi-scene multi-label emotional dialogue database. In: Muresan, S., Nakov, P., Villavicencio, A. (eds.) Proceedings of the 60th Annual Meeting of the Association for Computational Linguistics (Volume 1: Long Papers). pp. 5699--5710. Association for Computational Linguistics, Dublin, Ireland (May 2022). \doi{10.18653/v1/2022.acl-long.391}, \url{https://aclanthology.org/2022.acl-long.391}

\bibitem{minigpt5}
Zheng, K., He, X., Wang, X.E.: Minigpt-5: Interleaved vision-and-language generation via generative vokens. arXiv preprint arXiv:2310.02239  (2023)

\bibitem{related_mmchat}
Zheng, Y., Chen, G., Liu, X., Sun, J.: {MMC}hat: Multi-modal chat dataset on social media. In: Calzolari, N., B{\'e}chet, F., Blache, P., Choukri, K., Cieri, C., Declerck, T., Goggi, S., Isahara, H., Maegaard, B., Mariani, J., Mazo, H., Odijk, J., Piperidis, S. (eds.) Proceedings of the Thirteenth Language Resources and Evaluation Conference. pp. 5778--5786. European Language Resources Association, Marseille, France (Jun 2022), \url{https://aclanthology.org/2022.lrec-1.621}

\end{thebibliography}

\clearpage
\appendix
\section{Limitations}
\label{appendix:limtations}
Our framework relies on customized text-to-image models to ensure image consistency in multimodal dialogues. While these models generally offer better consistency than standard text-to-image models without conditioning, they are not infallible and may sometimes fail to accurately capture the conditioned input image. This represents a current limitation of our work. However, with the rapid advancements in customized text-to-image generation, we expect these shortcomings to decrease over time. 

\section{Broader Impact}
\label{appendix:braoder_impact}
It is crucial to emphasize that the main contribution of our work is not the customized text-to-image model itself but the overall framework that facilitates its effective use in multimodal dialogue scenarios. By focusing on enhancing image consistency, our framework opens up new avenues for more coherent and engaging multimodal interactions. This underscores the potential of our approach in revolutionizing how conversational agents handle multimodal inputs and responses, paving the way for more sophisticated and human-like dialogue systems.

\section{Benchmark Datasets}
\subsection{Categorization of Existing Multimodal Dialogue Datasets.}
\label{appendix:category_dataset}
As stated in Section \ref{related_works}, Multimodal dialogue datasets generally fall into three categories: question and answering (Q\&A), in-scene, and conversation-based. In Table \ref{tab:dataset_category}, we summarize the datasets for each category.

\begin{table}[h!]
\centering
\scriptsize
\caption{Summary of Multimodal Dialogue Datasets. The type can generally be classified into three categories: question and answering (Q\&A), the conversation taking place in a scene from a video (in-scene), and natural multimodal conversation (conversation-based). The modalities can contain audio (a), video (v), image (i), or text (t).}
\label{tab:dataset_category}
\begin{tabular}{@{}cccccccc@{}}
\toprule
\textbf{Dataset} & \textbf{Dialogue Type} & \textbf{Modalities} & \textbf{Dialogue Source} & \textbf{Turns} & \textbf{Language} & \textbf{Public}  \\ \midrule
VisDial \cite{related_visdialog} & Q\&A & i,t & crowd-sourcing & 2.47M & English & o\\
AVSD \cite{related_avsd} & Q\&A & a,v,t & crowd-sourcing & 236K & English & o\\
\midrule
OpenViDial \cite{related_openvidial} & in-scene & i,t & movies\&TVs & 1.1M & English & o\\
OpenViDial 2.0 \cite{related_openvidial2} & in-scene & i,t & movies\&TVs & 5.6M & English & o\\
YTD-18M \cite{related_ytd} & in-scene & a,v,t & movies\&TVs & 5.6M & English & o\\
\midrule
ImageChat \cite{imagechat} & conversation-based & i,t & crowd-sourcing & 401K & English & o\\
PhotoChat \cite{related_photochat} & conversation-based & i,t & crowd-sourcing & 156K & English & o\\
MMDD \cite{related_mmdd} & conversation-based & i,t & text datasets & 346K & English & o\\
DialogCC \cite{related_dialogcc} & conversation-based & i,t & text datasets & 929K & English & x\\
MMDialog \cite{mmdialog} & conversation-based & i,t & social media & 4.92M & English & o \\
MMChat \cite{related_mmchat} & conversation-based & i,t & social media & 314K & Chinese & o\\
TikTalk \cite{related_tiktalk} & conversation-based & a,v,t & social media & 827K & Chinese & o\\

\bottomrule
\end{tabular}
\end{table}
\begin{figure}[t]
	\centering
	\includegraphics[width=1.0\linewidth]{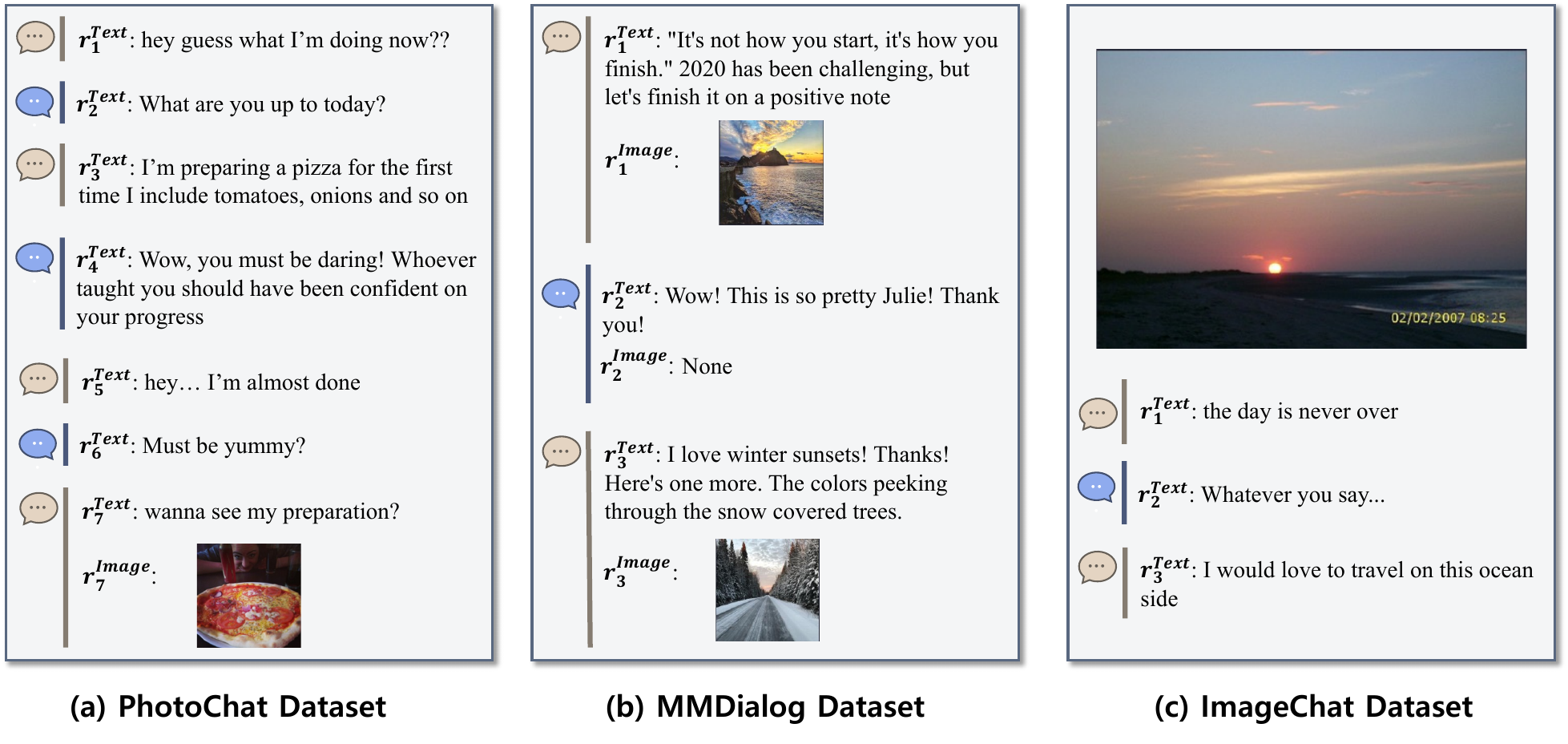}
    \caption{\textbf{Example of Benchmark Datasets used in our paper.}}
	\label{fig:dataset_example}
\vspace{5pt}
\end{figure}

\subsection{PhotoChat Dataset}
\label{appendix:photochat_example}
PhotoChat \cite{related_photochat} features dialogues collected from social media, where a single image is shared in one of the conversation turns, which mirrors everyday human interaction. 
An example of PhotoChat dialogue is shown in Figure \ref{fig:dataset_example}-(a).

\subsection{MMDialog Dataset}
\label{appendix:mmdialog_example}
The limited scale and domain diversity of the PhotoChat dataset restricts its applicability. Overcoming these limitations, MMDialog \cite{mmdialog} features over a million diverse dialogues from social media, where multiple images are shared across numerous conversation turns, providing a more realistic representation of open-domain multimodal conversations. An example of MMDialog dialogue is shown in Figure \ref{fig:dataset_example}-(b).

\subsection{ImageChat Dataset}
\label{appendix:imagechat_example}
To evaluate the image-grounding advantage of our BI-MDRG to the previous system, we use the ImageChat Dataset \cite{imagechat}. This dataset has three turns of conversation about a given image. 
An example of ImageChat Dialogue is shown in Figure \ref{fig:dataset_example}-(c).

\subsection{Multimodal Dialogue Image Consistency (MDIC) Dataset}
\label{appendix:mdic}
The challenge of ensuring consistent image generation in multimodal dialogue systems is amplified by the absence of datasets annotated for entity consistency across conversational images. We developed the Citation Module for our BI-MDRG system to address this gap. This module is designed to pseudo-label the recurring visual entities throughout a dialogue, allowing us to train our model to generate textual image descriptions during inference with citations that reflect the objects needing consistency. However, a benchmark dataset with explicit image consistency annotation is essential to validate the Citation Module and our BI-MDRG, which was trained with the pseudo-labels created from the Citation Module. To this end, we created the Multimodal Dialogue Image Consistency (MDIC) dataset. This dataset comprises a collection of dialogues annotated to identify the recurring visual entities across the conversation. 
\begin{figure}[t!]
	\centering
	\includegraphics[width=1.0\linewidth]{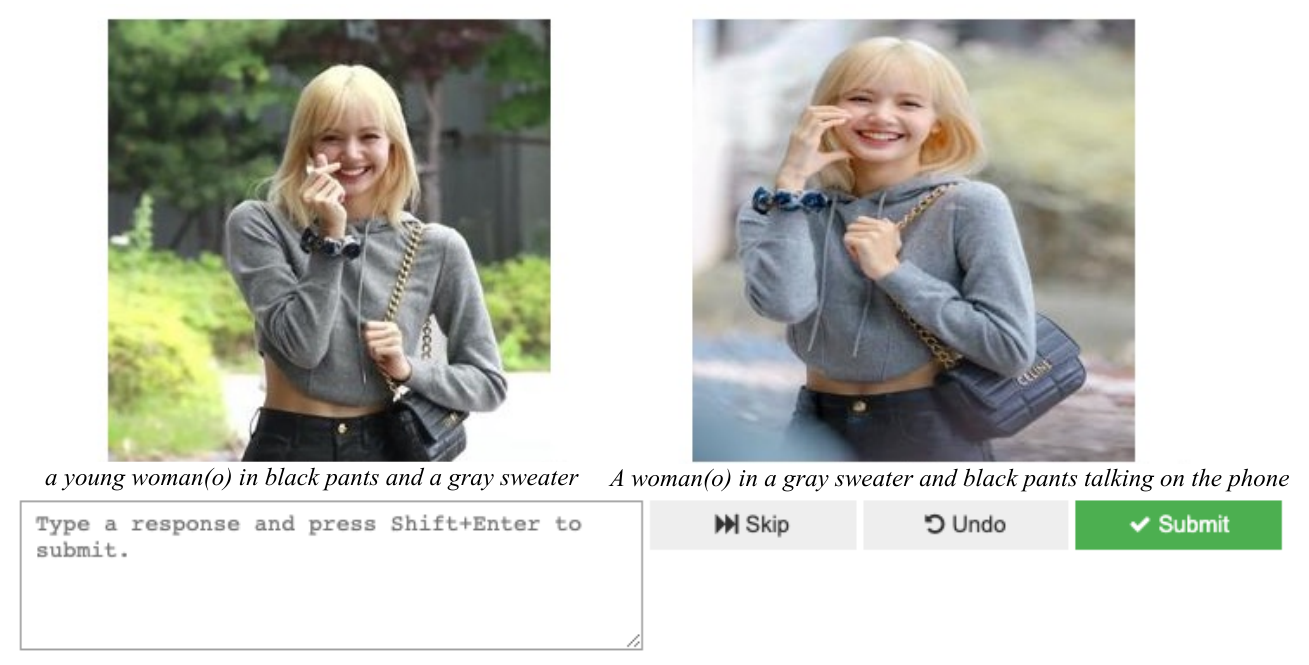}
	\caption{Illustration of the labeling interface used for creating the Multimodal Dialogue Image Consistency (MDIC) dataset. The interface presents all images associated with a specific dialogue from the MMDialog test set. Labelers are tasked with assigning citation tags to the primary objects in these images, identified as \textit{(o)}. The assignment is based on visual similarity and the identity of objects across different images.}
	\label{fig:labeling_process}
\end{figure}

\subsubsection{Labeling Process} 
\begin{figure}[htb]
	\centering
	\includegraphics[width=0.80\linewidth]{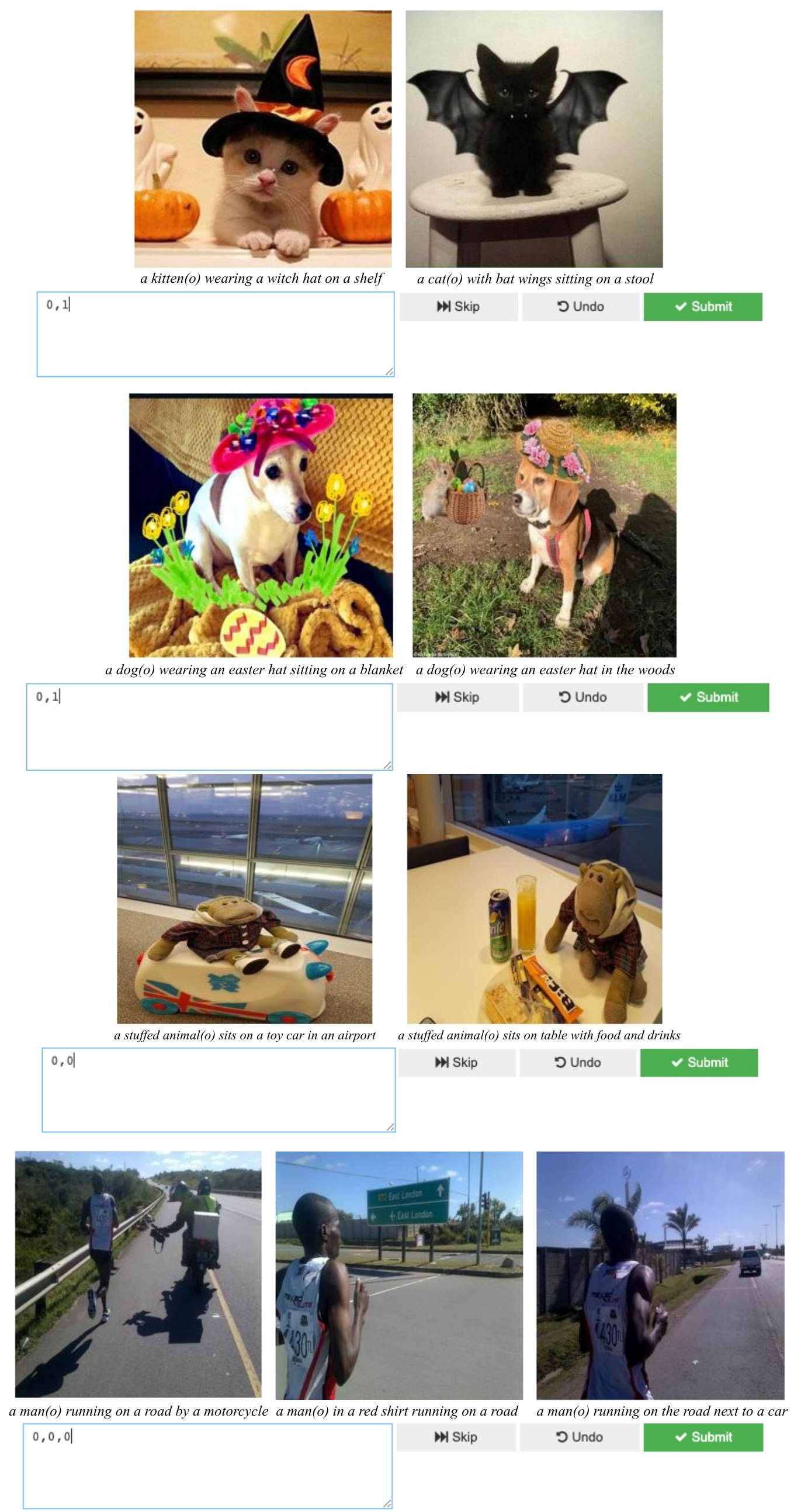}
	\caption{Examples of labeled annotations of the MDIC dataset. The labeler inputs comma-separated numbers that represent the citation of the primary object in the textual image description based on the object's similarity.}
	\label{fig:labeling_examples}
\end{figure}

MDIC benchmark dataset was created using a labeling process applied to the images from the MMDialog test set. Figure \ref{fig:labeling_process} illustrates the labeling interface used. For each dialogue's images, its corresponding textual image descriptions were obtained using BLIP2-flan-t5-xl \cite{blip2} and pre-processed using spaCy \cite{spacy2} to identify the primary objects in the sentence. Five annotators examined these images and descriptions and assigned citation tags to the primary objects based on visual similarity and the identity of the objects across different images (examples of annotations are shown in Figure \ref{fig:labeling_examples}). For instance, if a dialogue contained two images with the same object, the labeler would input `0,0'; if the two images contained different objects, the labeler would input `0,1'. The final dataset selections were based on a consensus approach, retaining only those test sets where all five annotators unanimously agreed. 
\clearpage
\clearpage
\begin{figure}[t!]
	\centering
	\includegraphics[width=0.90\linewidth]{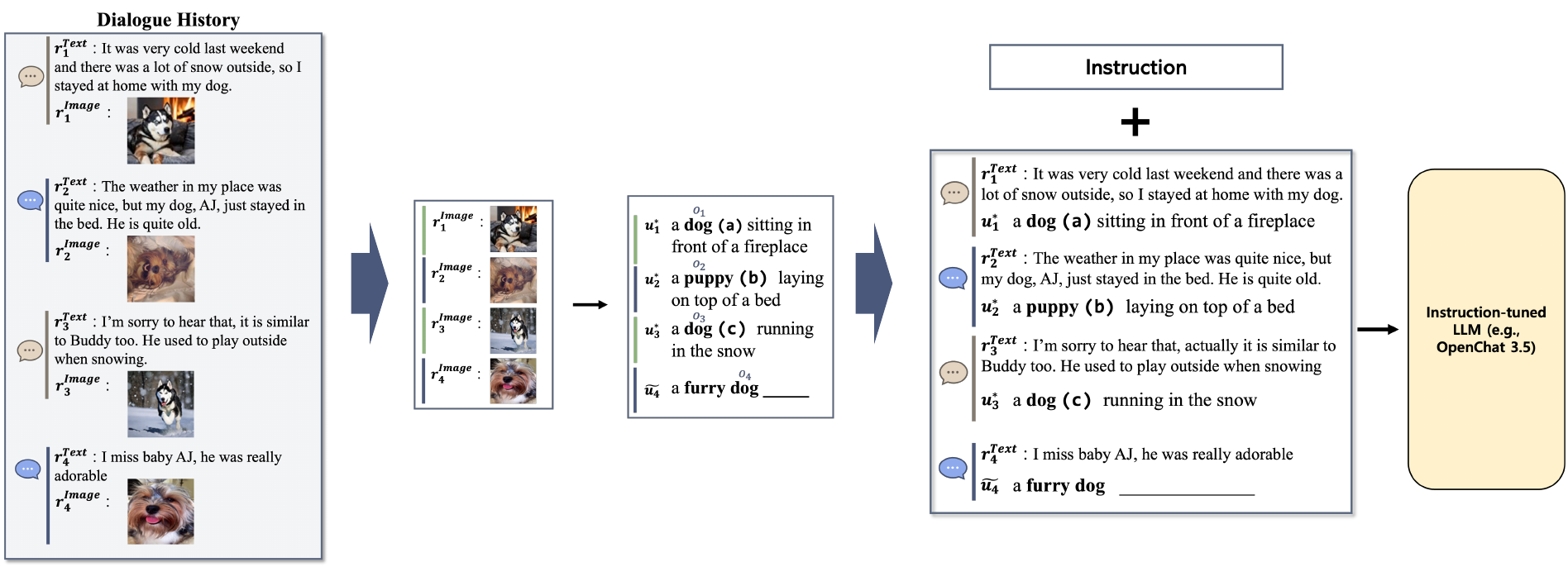}
	\caption{\textbf{Illustration of the LLMCite Baseline.} This approach employs an instruction-tuned large language model for assigning citation tags, treating citation tag prediction as a multiclass classification task. Specifically, it involves selecting the object from the dialogue history that the current object identically matches.}
	\label{fig:LLMCite}
 \vspace{30pt}
\end{figure}

\begin{figure}[t!]
	\centering
	\includegraphics[width=1.0\linewidth]{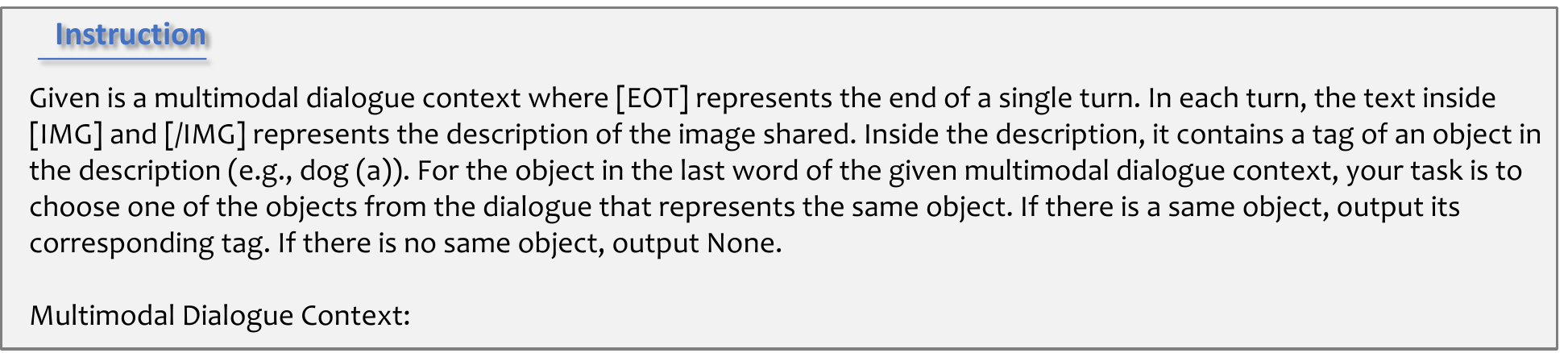}
	\caption{\textbf{Instruction given to the LLM for the LLMCite baseline.}}
	\label{fig:instruction}
 \vspace{30pt}
\end{figure}

\section{Details on LLMCite}
In Sections \ref{sec4.4} and \ref{section5}, we employ a baseline citation approach, LLMCite, illustrated in Figure \ref{fig:LLMCite}, which leverages an instruction-tuned large language model (LLM) to assign citation tags (specifically, we use \textit{OpenChat 3.5 (7B)}\footnote{\url{https://huggingface.co/openchat/openchat\_3.5}}). From the MDIC dataset, we frame citation tag prediction as a multiclass classification task. Given a dialogue history $D=\{(r_i^{\text{{Text}}}, r_i^{\text{{Image}}})\}_{i=1}^t$, we first convert images into textual descriptions to form $\{r_1^{\text{{Text}}},u_1, \ldots, r_{t}^{\text{{Text}}},u_t\}$. For the last turn $t$, we preprocess $u_t$ to include only up to the principal object $o_t$, denoted as $\tilde{u_t}$. For preceding turns $u_{1:t-1}$, we append classification tags $c^*_{1:t-1}$ (sequentially labeled as (a), (b), (c), ...) to principal objects $o_{1:t-1}$, resulting in augmented descriptions $u^*_{1:t-1}$. This modified sequence $\{r_1^{\text{{Text}}},u^*_1, \ldots, r_{t}^{\text{{Text}}},\tilde{u_t}\}$ is then provided to the LLM with instructions, as illustrated in Figure \ref{fig:instruction}, to choose the most appropriate $c^*_{1:t-1}$ matching $o_t$ within the dialogue context.

\label{appendix:LLMcite}

\section{Additional Examples}
\label{appendix: additional_examples}
\begin{figure}[h!]
	\centering
	\includegraphics[width=1.0\linewidth]{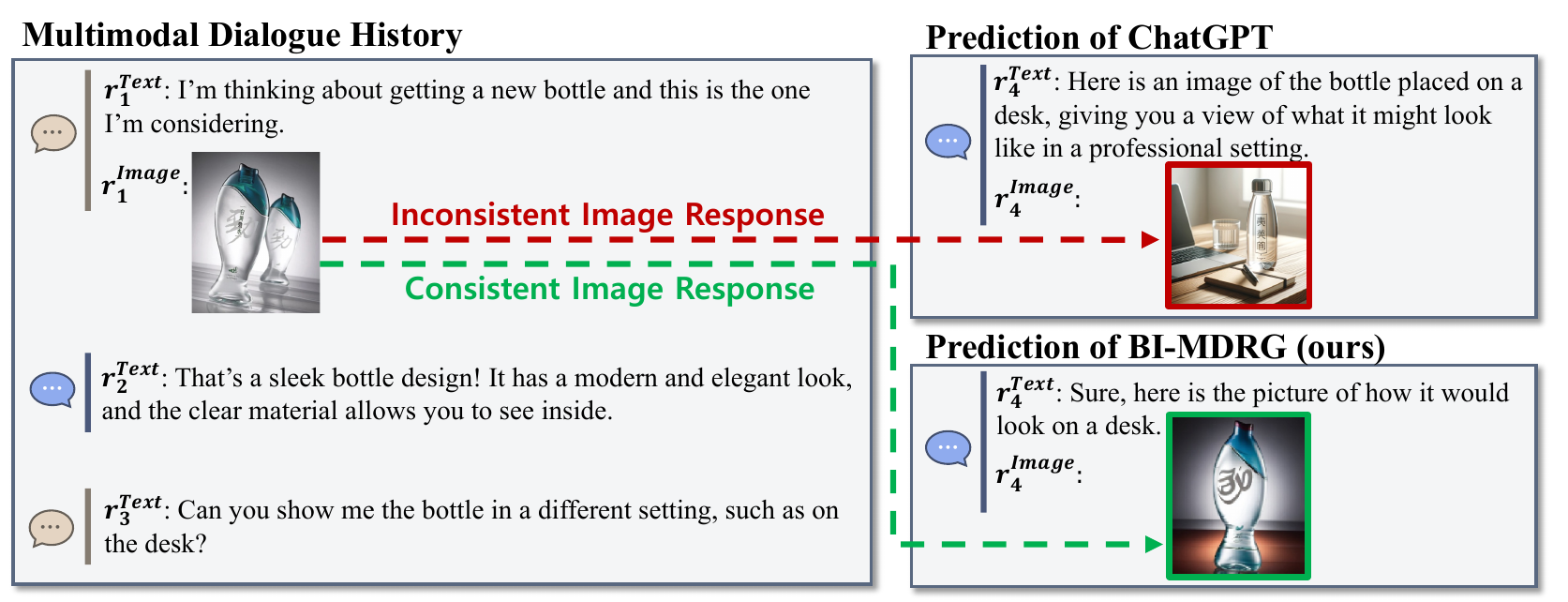}
	\caption{\textbf{Image Response of ChatGPT and BI-MDRG (ours).}}
	\label{fig:chatpgt}
\end{figure}

In Section \ref{section5}, we demonstrated that merely increasing the model size does not enhance image consistency. This limitation arises because the framework relies on text as an intermediary step for generating image responses, leading to an inherent loss of image information. ChatGPT also operates within this framework, utilizing text as an intermediary due to the challenges and infeasibility of implementing an end-to-end model, a point underscored in Section \ref{section5}. Consequently, our proposed framework, specifically designed to maintain image consistency, becomes critical. Figure \ref{fig:chatpgt} illustrates that ChatGPT also struggles to maintain image consistency, reinforcing the need for our targeted framework.

\begin{figure}[h]
	\centering
 \vspace{-7pt}
	\includegraphics[width=1.0\linewidth]{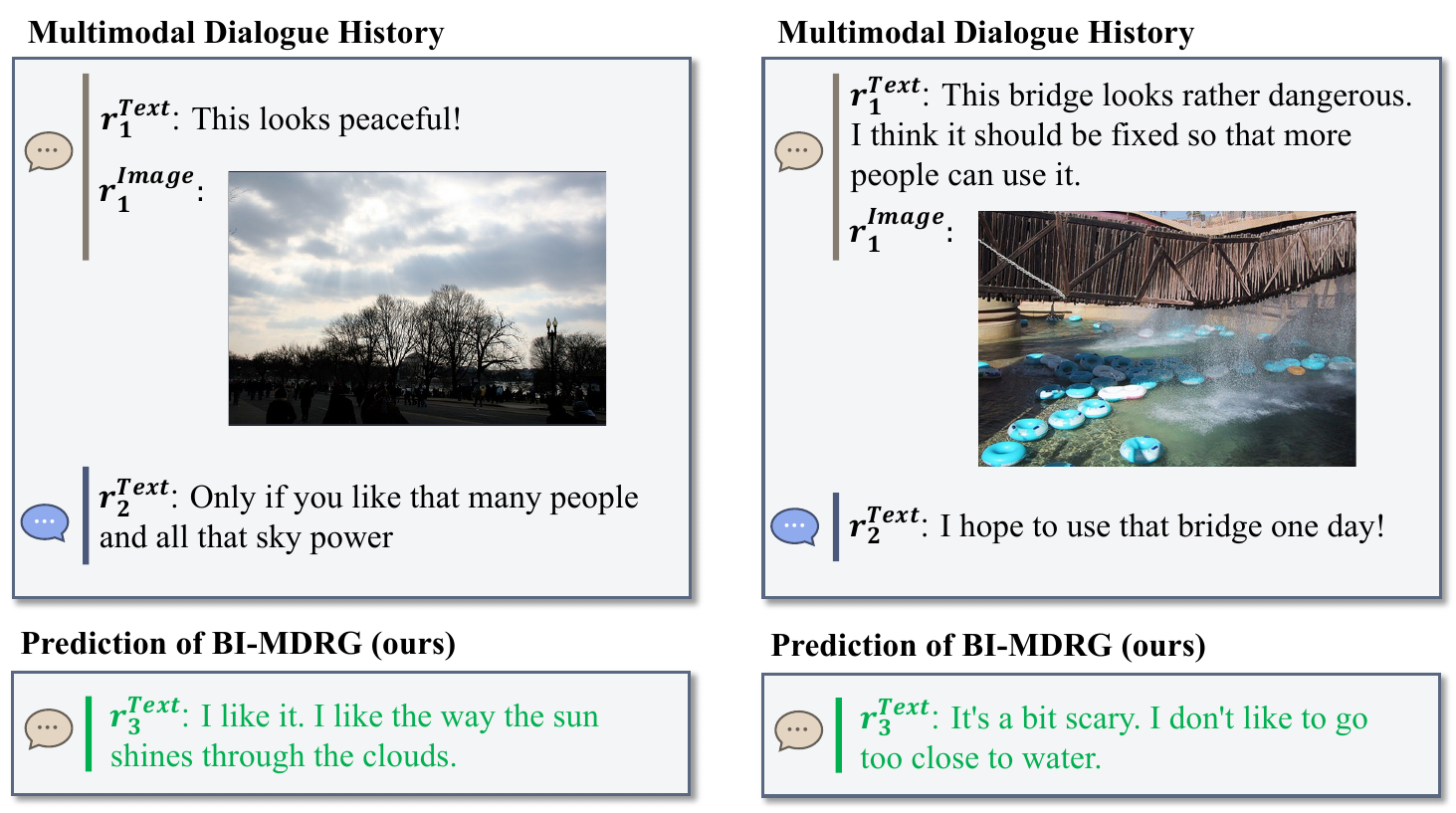}
	\caption{\textbf{Examples of Image-Grounded Text Response of BI-MDRG (ours).}}
	\label{fig:imagegrounding_extra}
\end{figure}

\begin{figure}[h]
	\centering
	\includegraphics[width=1.0\linewidth]{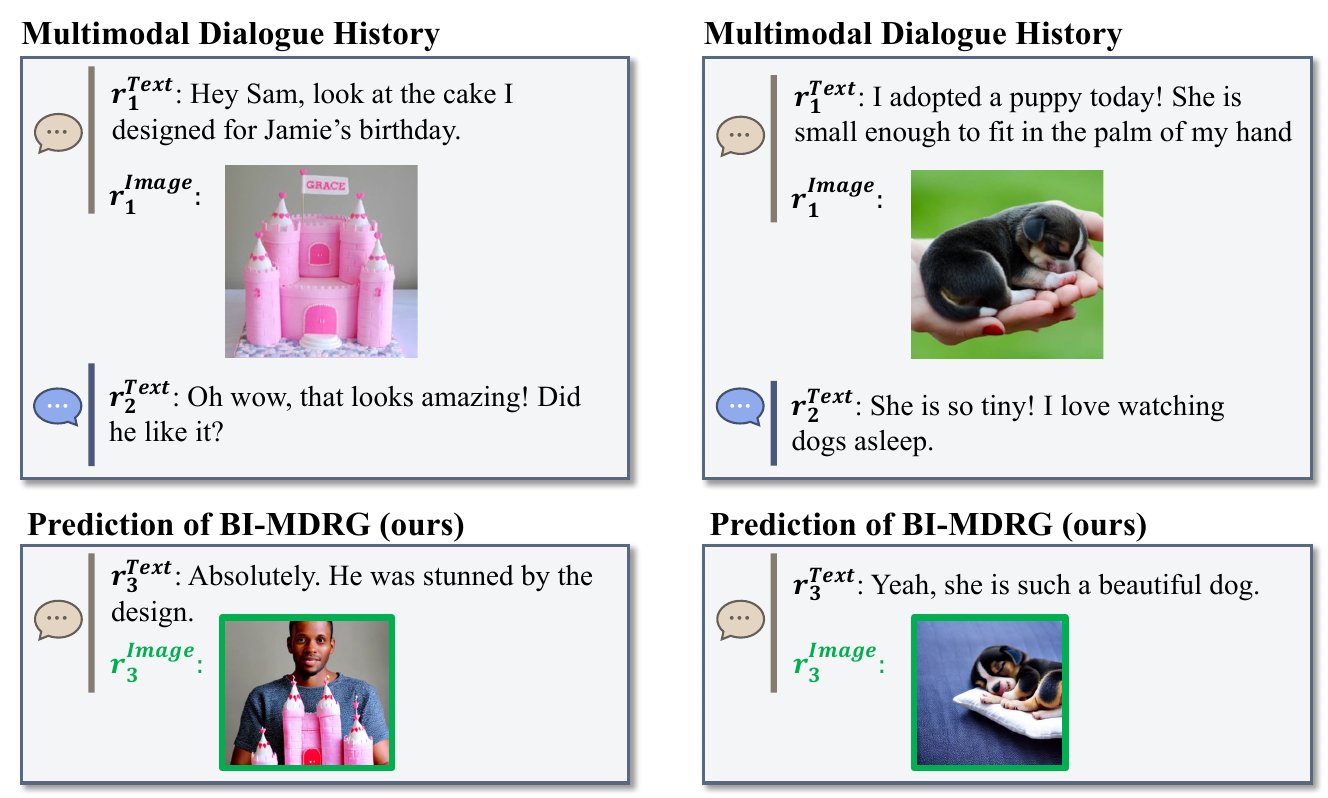}
	\caption{\textbf{Examples of Consistent Image Response of BI-MDRG (ours).}}
	\label{fig:imageconsistency_extra}
\end{figure}

In Figures \ref{fig:imagegrounding_extra} and \ref{fig:imageconsistency_extra}, we present further examples of BI-MDRG predictions, showcasing both image-grounded textual responses and the model's ability to maintain consistency in image responses.


\end{document}